\documentclass[a4paper,fleqn]{cas-sc}

\usepackage[authoryear,longnamesfirst]{natbib}
\usepackage{booktabs}
\usepackage{multirow}
\usepackage[normalem]{ulem} 

\def\tsc#1{\csdef{#1}{\textsc{\lowercase{#1}}\xspace}}
\tsc{WGM}
\tsc{QE}

\begin{document}
\let\WriteBookmarks\relax
\def\floatpagepagefraction{1}
\def\textpagefraction{.001}
\let\printorcid\relax

\shorttitle{TunnelMIND}

\shortauthors{S. Liu et~al.}  

\title [mode = title]{Training-Free Tunnel Defect Inspection and Engineering Interpretation via Visual Recalibration and Entity Reconstruction}

\author[1]{Shipeng Liu}
\ead{lsp@xauat.edu.cn}
\affiliation[1]{organization={Xi'an University of Architecture and Technology},
            department={School of mechanical and electrical engineering,},
            city={Xi'an},
            postcode={710055}, 
            state={Shaanxi},
            country={China}}

\author[2]{Liang Zhao}
\ead{zhaoliang@xauat.edu.cn}
\affiliation[2]{organization={Xi'an University of Architecture and Technology},
            department={School of Information and Control Engineering,},
            city={Xi'an},
            postcode={710055}, 
            state={Shaanxi},
            country={China}}

\author[3]{Dengfeng Chen}
\ead{chdengf@xauat.edu.cn}
\affiliation[2]{organization={Xi'an University of Architecture and Technology},
            department={School of Information and Control Engineering,},
            city={Xi'an},
            postcode={710055}, 
            state={Shaanxi},
            country={China}}

\author[4]{Zhanping Song}
\cormark[1]
\ead{songzhpyt@xauat.edu.cn}
\affiliation[3]{organization={Xi'an University of Architecture and Technology},
            department={School of Civil Engineering,},
            city={Xi'an},
            postcode={710055}, 
            state={Shaanxi},
            country={China}}

\cortext[1]{Corresponding author}

\begin{abstract}
Tunnel inspection requires outputs that can support defect localization, measurement, severity grading, and engineering documentation. Existing training-free foundation-model pipelines usually stop at coarse open-vocabulary proposals, which are difficult to use directly in interference-heavy tunnel scenes. We propose  a training-free framework TunnelMIND. Specifically, language-guided defect proposals are not treated as final outputs; instead, their spatial support is recalibrated at inference time through dense visual consistency, so that coarse semantic anchors can be transformed into more reliable prompts under tunnel-specific hard negatives. The resulting masks are further reconstructed into structured defect entities with category, location, geometry, severity, and context attributes, which are then mapped to retrieval-grounded explanation and engineering-readable report generation under expert knowledge constraints. On visible, GPR, and road defect tasks, TunnelMIND achieves F1 scores of 0.68, 0.78, and 0.72, respectively. Overall, TunnelMIND shows that training-free tunnel inspection can move beyond coarse localization toward structured defect evidence for engineering assessment.
\end{abstract}




\maketitle


\section{Introduction}\label{sec1:intro}
Tunnel inspection requires more than category-level recognition. In practical engineering workflows, inspection outputs are expected to indicate what defect may be present, where it is located, how large it is, and whether its severity is sufficient to require further attention. This requirement makes tunnel inspection fundamentally different from generic vision tasks. Even when a model can indicate potentially relevant regions, its outputs may still be unsuitable for engineering use if the spatial support is unstable, the boundaries are unreliable, or the results cannot be converted into measurable and traceable defect records for subsequent evaluation, maintenance planning, and inspection documentation \citet{cho2026bim,ou2025towards,attard2017image}.

Such requirements arise throughout the tunnel engineering lifecycle. During construction, visual perception is needed for tasks such as excavation-face observation, rock slag assessment, worker safety monitoring, and equipment compliance checking \citet{shi2014advance, vukicevic2022generic}. During operation and maintenance, visible lining defects, hidden defects revealed by ground-penetrating radar (GPR), and road-surface damage must be detected and interpreted in a form that supports condition assessment and follow-up intervention \citet{ou2025towards}. Across these scenarios, tunnel images are frequently affected by dense hard negatives, including shadows, structural joints, stains, reflections, and repetitive textures. These disturbances do not merely reduce visual clarity. More importantly, they make it difficult to distinguish true defects from visually similar background responses, which directly limits the reliability of engineering quantification and record generation \citet{sjolander2023towards,zhang2024reactive}.

Recent supervised deep learning methods have achieved strong performance on tunnel crack detection, seepage segmentation, GPR defect interpretation, pavement damage detection, and safety-related monitoring when task-specific annotations are sufficient and deployment conditions remain relatively stable \citet{zhou2024tunnel,feng2023deep,zhao2025multimodal}. However, three practical limitations still remain. First, collecting and finely annotating tunnel data is expensive, especially when different tunnel sections, construction stages, sensing modalities, and imaging conditions must all be covered. Second, the resulting models are often tightly coupled to the training distribution, making transfer to new sites, new tasks, or new acquisition conditions difficult without further retraining. Third, the one-model-per-task paradigm is increasingly difficult to maintain in engineering deployment, where inspection needs evolve across construction and operation stages but still require a coherent and reusable technical interface \citet{su2024review}.

Foundation models and multimodal large models provide a promising alternative to this conventional workflow. Promptable segmentation models \citet{kirillov2023segment}, self-supervised dense visual encoders, and vision-language models have demonstrated strong transferability under limited or even zero task-specific optimization. For tunnel engineering, this direction is attractive because it may reduce the burden of collecting task-specific annotations while increasing deployment flexibility across tasks. However, current training-free pipelines often stop at coarse open-vocabulary grounding \citet{zhang2024vision,myers2024foundation}. In tunnel scenes, this is not enough. A language-guided model may return a plausible defect category or a roughly relevant region, yet still fail to provide the spatial reliability needed for measurement, location assignment, severity grading, and engineering documentation. In other words, the key bottleneck is not merely whether a model can find suspicious regions, but whether coarse proposals can be converted into stable, measurable, and traceable defect entities for engineering interpretation.

\begin{figure}
    \centering
    \includegraphics[width=0.9\linewidth]{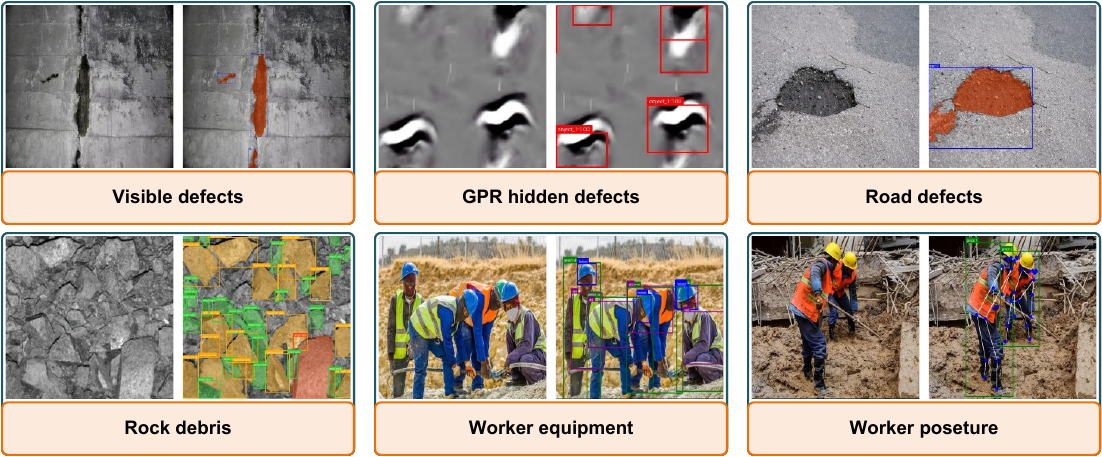}
    \caption{\textbf{Representative tunnel inspection tasks considered in this work}, covering both construction and operation scenarios. In each task group, the left image is the original input and the right image shows the corresponding visualization result. From left to right and top to bottom, the examples include blasting rock mass, tunnel lining disease, tunnel GPR defects, pavement disease, worker equipment, and worker status. The figure highlights the substantial variation in appearance, modality, and engineering semantics across tunnel scenes.}
    \label{fig:1}
    \vspace{-10pt}
\end{figure}

This bottleneck becomes especially important in interference-heavy tunnel environments. Shadows may resemble cracks or seepage boundaries, construction joints may be confused with structural defects, stains and contamination may trigger false positives, and repetitive tunnel textures may cause proposal drift or duplication. Similar difficulties also arise in weak-semantic modalities such as GPR, where abnormal responses are visually meaningful but not always naturally aligned with explicit language descriptions. As a result, directly treating the output of a vision-language model as the final inspection result is often insufficient for engineering use. What is needed is an intermediate mechanism that can preserve the semantic flexibility of training-free perception while improving the spatial reliability and engineering interpretability of the resulting outputs.

To address this gap, we propose TunnelMIND \footnote{TunnelMIND is the name of the proposed framework, reflecting its focus on training-free tunnel inspection, entity-centric interpretation, and retrieval-grounded assistance.}, a training-free tunnel inspection framework that bridges language-guided semantic proposals and engineering-ready inspection entities. Instead of using a vision-language model as the final detector, TunnelMIND treats it as a semantic anchor generator. These anchors are then transferred into the dense feature space of DINOv3, where local visual consistency is used to recalibrate spatial support under tunnel-specific hard negatives \citet{xu2024training,bai2023qwen,simeoni2025dinov3}. The recalibrated support is further converted by SAM into pixel-level masks, from which structured engineering entities are reconstructed with category, location, geometry, severity, and context attributes. (Figure~\ref{fig:1} presents the six tunnel visual perception tasks selected in this work and their corresponding visual results.) Based on these entities, the framework can further support retrieval-grounded explanation and engineering-readable report assistance under explicit knowledge constraints \citet{gao2023retrieval,lewis2020retrieval}.

Different from existing zero-shot or training-free inspection frameworks that mainly target single tasks, relatively simple backgrounds, or box/mask-level outputs, TunnelMIND is developed for interference-heavy tunnel scenarios in which engineering usability is the central requirement. The framework is evaluated on six tunnel-related tasks spanning both construction and operation settings, including rock debris segmentation, visible defect detection, GPR hidden defect detection, road defect detection, PPE inspection, and worker pose recognition. Among them, visible defects, GPR defects, and road defects are used as representative defect-oriented tasks for the main analysis because they directly reflect the challenges of weak boundaries, irregular geometry, dense interference, and engineering quantification. The remaining tasks are used to examine interface-level transferability.

The main contributions of this work are summarized as follows:

\begin{itemize}
    \item We formulate tunnel inspection as a training-free engineering interpretation problem, where the central challenge is not only defect indication but also the transformation of coarse proposals into measurable and traceable defect entities.
    \item We propose an inference-time cross-model visual recalibration strategy that transfers language-guided semantic anchors into a dense visual support space and improves spatial reliability under tunnel-specific hard negatives.
    \item We introduce an entity-centric engineering reconstruction mechanism that converts recalibrated masks into structured instances with category, location, geometry, severity, and context attributes, making the outputs more compatible with engineering quantification and inspection records.
    \item We validate the framework on representative tunnel defect tasks as well as additional tunnel-related tasks, and further show that the resulting structured entities provide a more effective interface for retrieval-grounded explanation and engineering-readable report assistance than direct large-model outputs.
\end{itemize}

The remainder of this paper is organized as follows. Section~\ref{sec2:related work} reviews related work on tunnel visual perception, foundation-model-based inspection, and retrieval-grounded engineering assistance. Section~\ref{sec3:method} presents the proposed TunnelMIND framework, including semantic anchoring, cross-model visual recalibration, and entity-centric engineering reconstruction. Section~\ref{sec4} reports experimental results on representative tasks, hard-negative subsets, cross-task transferability, entity-level engineering evaluation, and reporting utility. Section~\ref{sec5} discusses engineering implications, limitations, and future directions. Section~\ref{sec6} concludes the paper.

\section{Related work}\label{sec2:related work}
\subsection{Tunnel visual perception}
Tunnel construction and maintenance involve a wide range of visual perception tasks, including excavation-face analysis, rock slag recognition, PPE compliance monitoring, worker action understanding, and the detection of visible defects such as cracks, seepage, spalling, and pavement damage during operation \citet{vukicevic2022generic,yu2024road}. Recent reviews have shown that optical tunnel inspection must jointly consider defect observability, imaging conditions, and assessment-oriented data acquisition, rather than treating tunnel defects as isolated recognition targets \citet{sjolander2023towards}. Most existing studies formulate these problems as task-specific supervised learning tasks and address them with dedicated detection, segmentation, or instance segmentation models \citet{zhao2020deep,zhou2024tunnel,feng2023deep}. For example, visible defect detection is commonly tackled with YOLO \citet{redmon2016you}, DETR \citet{carion2020end}, or related detector-based frameworks, while region-oriented targets such as cracks and seepage are often handled by U-Net \citet{ronneberger2015u} variants with attention, feature fusion, or lightweight designs. For hidden tunnel defects, ground-penetrating radar (GPR) interpretation has also gradually shifted from manual inspection to deep learning-based automatic recognition and segmentation \citet{zhao2025multimodal}.

These approaches can achieve strong performance when annotations are sufficient, scene conditions are controlled, and task boundaries are clearly defined. However, their effectiveness typically relies on abundant labeled data and relatively stable deployment environments. In tunnel engineering, this assumption is difficult to maintain over time because geological conditions, lighting arrangements, construction methods, and camera viewpoints vary substantially across sites and stages. In addition, decomposing multi-task perception into multiple isolated models leads to increasing deployment, maintenance, and upgrade costs. As a result, existing supervised tunnel inspection methods are more suitable for single-task optimization than for cross-stage, cross-site, and continuously evolving engineering inspection demands \citet{su2024review,zhang2024reactive}.

\subsection{Foundation models for perception}
The development of foundation models has opened a new path toward reducing dependence on task-specific training \citet{myers2024foundation}. SAM reformulates segmentation as a promptable task, allowing points and boxes to serve as transferable interfaces across categories and data distributions \citet{kirillov2023segment}. Beyond generic promptable segmentation, subsequent studies such as SEEM and Semantic-SAM further extended the idea toward more universal and semantically aware segmentation interfaces, broadening the range of prompt types and granularity levels that can be handled within a unified framework \citet{zou2023segment,li2024segment}. Self-supervised visual representation models provide dense features with strong transferability, which can support region matching, similarity retrieval, and inference-time constraints without explicit task supervision \citet{chen2023mixed,oquab2023dinov2}. Earlier unsupervised discovery methods such as LOST and TokenCut further showed that self-supervised transformer features can already support label-free localization and region grouping through feature consistency alone, even without task-specific training \citet{simeoni2021localizing,wang2022self}. At the same time, open-vocabulary detection introduces language conditions into visual localization, enabling models to search for image regions beyond a fixed label space \citet{zhu2024survey}. Representative paradigms in this direction include grounded pretraining methods such as GLIP, region-level vision-language alignment methods such as RegionCLIP, scalable detector adaptations such as OWL-ViT, and unified open-vocabulary segmentation frameworks such as OpenSeeD \citet{li2022grounded,minderer2022simple,zhong2022regionclip,zhang2023simple}.

These advances suggest that training-light, or even training-free, perception is becoming feasible. However, two limitations remain critical in tunnel environments. First, tunnel imagery contains dense hard negatives such as joints, shadows, stains, and repetitive textures, so language-aligned open-vocabulary proposals are prone to false positives, missed detections, and duplicates. Second, engineering applications require not only semantic discovery but also stable spatial boundaries and measurable geometric attributes to support downstream analysis such as length, area, location, and severity estimation. Current foundation models therefore provide the potential for open-vocabulary perception and prompt-based segmentation, but still lack a reliable intermediate step that converts semantic proposals into engineering-ready entities.

\subsection{Vision-language models for inspection}
Vision-language models and multimodal large models further lower the barrier to building complex visual systems \citet{wu2023multimodal,zhang2024vision}. Compared with conventional task-specific models, these methods use natural language as a unified interface and can generate candidate regions, coordinates, or structured responses from category phrases or task descriptions. For tunnel scenarios, where tasks are diverse, requirements change frequently, and annotations are expensive, such language-driven candidate generation is particularly appealing \citet{yang2025qwen3}.

However, a unified interface does not automatically imply engineering reliability. In tunnel scenes, the main challenge is often not whether a multimodal model can produce some candidate response, but whether the response has sufficiently stable spatial support under hard-negative, low-contrast, and texture-interference conditions. Language models may provide a semantic answer to what is present, but they do not necessarily provide a reliable answer to where it is and whether its boundary is trustworthy enough for engineering quantification. Therefore, an engineering-oriented inspection system should not stop at language-guided candidate generation, but should further combine semantic anchors with dense visual consistency to recalibrate and constrain open-vocabulary proposals at inference time. Most existing vision-language pipelines still treat semantic grounding as an almost direct proxy for usable localization, whereas engineering scenarios often require an additional stage that converts semantic cues into spatially more stable support before geometric measurement.

\subsection{Retrieval-grounded report assistance}
Beyond perception itself, engineering deployment also requires model outputs to be converted into inspection explanations, risk interpretation, and engineering-readable reporting. The core idea of retrieval-augmented generation is to couple parametric generation with explicit non-parametric memory, thereby improving factual grounding, provenance, and updateability in knowledge-intensive tasks \citet{lewis2020retrieval,gao2023retrieval}. In recent years, retrieval-grounded generation and knowledge-constrained reasoning have been increasingly used in domain-specific report assistance \citet{bao2025permitted}. Their basic idea is to map structured observations or language queries to retrieval requests, recall relevant evidence from standards, manuals, reports, and technical documents, and then generate constrained summaries and recommendations with the support of retrieved evidence. For infrastructure inspection and structural health monitoring, the value of such methods lies not in replacing professional engineers, but in improving traceability, consistency, and standard alignment during the explanation process.

\begin{figure}
    \centering
    \includegraphics[width=0.9\linewidth]{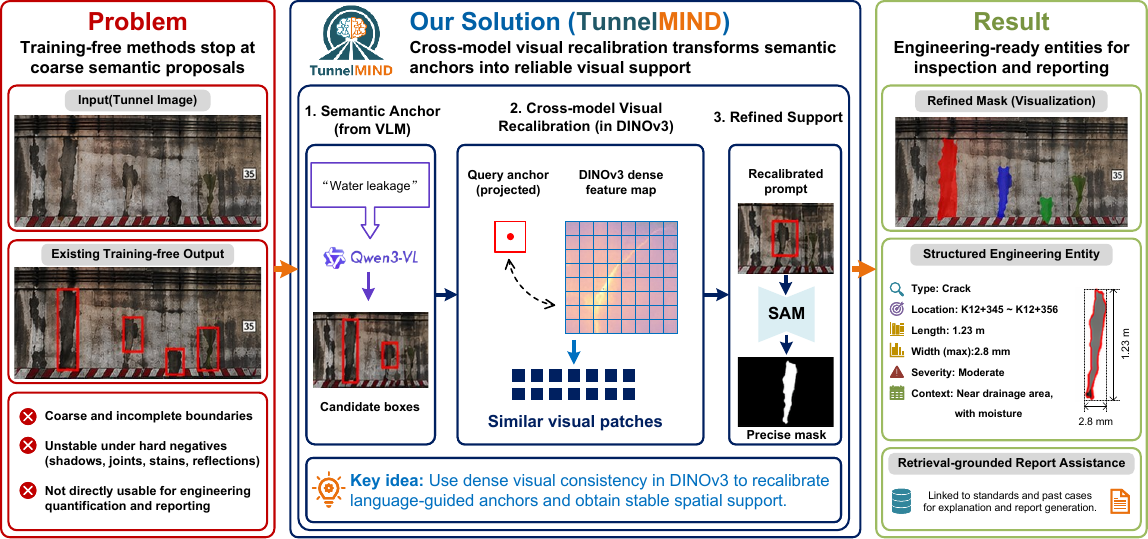}
    \caption{\textbf{Problem-to-solution overview of TunnelMIND.} Existing training-free tunnel inspection methods often stop at coarse semantic proposals. TunnelMIND bridges this gap by converting semantic anchors into reliable visual support, refined masks, and engineering-ready entities, thereby enabling spatially reliable, measurable, and traceable outputs for retrieval-grounded reporting.}
    \label{fig:gap}
\end{figure}

However, existing studies often focus on either visual recognition or knowledge-based question answering, with limited attention to the intermediate representation between them. If perception outputs remain at the level of boxes, masks, or free-form text, it is difficult to align them with geometric thresholds, severity levels, and intervention conditions in engineering documents. From the perspective of practical deployment, the key step is therefore not simply to feed visual results into a large language model (LLM) \citet{chang2024survey}, but to first convert perception outputs into structured engineering entities with category, location, geometry, and severity attributes, and then generate evidence-grounded explanations and advisory suggestions under retrieval constraints.

\subsection{Research gap}
In summary, prior work has advanced tunnel inspection from three related but still disconnected directions. Task-specific supervised models provide strong accuracy when annotations are sufficient and deployment conditions remain relatively stable. Foundation models and open-vocabulary methods enable training-light candidate generation and prompt-based segmentation. Retrieval-grounded reasoning further offers a promising interface for evidence-aware explanation and report generation. However, for tunnel inspection scenarios characterized by dense hard negatives, weak boundaries, and evolving task requirements, three gaps remain insufficiently addressed.

First, there is a spatial reliability gap: language-guided or open-vocabulary candidates in tunnel scenes often remain too unstable in their spatial support to be used directly for engineering quantification. Second, there is an entity reconstruction gap: current training-free outputs usually stop at boxes, masks, or free-form descriptions, while engineering workflows require structured entities with category, location, geometry, severity, and context attributes. Third, there is a perception-to-reporting gap: retrieval-grounded explanation methods typically assume that standardized structured observations are already available, but rarely address how such observations can be produced directly from raw tunnel imagery in a training-free manner. Figure~\ref{fig:gap} summarizes this problem-to-solution perspective. TunnelMIND is proposed precisely to fill this gap through cross-model visual recalibration and entity-centric engineering reconstruction.

TunnelMIND is developed to address these three gaps in a unified way by combining language-guided semantic anchoring, inference-time dense visual recalibration, prompt-based geometric realization, and entity-centric engineering reconstruction within a deployment-oriented training-free inspection pipeline.

\section{Method}\label{sec3:method}
\subsection{Problem formulation and overall pipeline}
Given an engineering image $I$ and a task-specific category set $C$, the goal of TunnelMIND is to produce measurable defect entities and retrieval-grounded reporting outputs without any task-specific parameter update. Rather than formulating tunnel inspection as a collection of isolated supervised detection or segmentation tasks, we cast it as a training-free inference framework that bridges the gap between open-vocabulary semantic cues and engineering-ready structured interpretation. Under this formulation, the key challenge is not merely whether a foundation model can roughly indicate potentially relevant regions, but whether its outputs can be transformed into spatially reliable, quantitatively measurable, and engineering-compatible entities under hard-negative tunnel conditions.

Rather than treating all components as equally weighted stages, we emphasize the core inference path from language-guided semantic proposals to engineering-ready entities. Formally, the central transformation in TunnelMIND can be written as
\begin{equation}
\widetilde{B}=\mathscr{R}\!\left(I,\Phi_{\mathrm{vlm}}(I,C)\right),
\qquad
E=\Gamma\!\left(\Psi_{\mathrm{sam}}(I,\widetilde{B}),\widetilde{B}\right)
\label{eq:core_pipeline}
\end{equation}
where $\Phi_{\mathrm{vlm}}$ denotes language-guided semantic anchoring, producing class-aware candidate proposals from the input image $I$ and task-specific category set $C$; $\mathscr{R}$ is the proposed cross-model visual recalibration module, which transforms these coarse semantic proposals into spatially more reliable support using dense DINOv3 features; $\Psi_{\mathrm{sam}}$ denotes prompted geometric realization that converts recalibrated support into pixel-level masks; and $\Gamma$ maps the resulting masks and proposal attributes to structured engineering entities $E$, including category, location, geometry, severity, and contextual information. In this formulation, the methodological core lies in $\mathscr{R}$ and $\Gamma$: $\mathscr{R}$ improves the spatial reliability of training-free semantic proposals under tunnel-specific hard negatives, while $\Gamma$ converts visually grounded masks into measurable and traceable engineering entities.

Based on the reconstructed entities, a deployment-facing reporting output can be further obtained as
\begin{equation}
y = G(E, K_n),
\label{eq:report_interface}
\end{equation}
where $G$ denotes retrieval-grounded explanation and report generation under expert knowledge constraints, and $K_n$ is the expert knowledge base. This reporting module is a downstream deployment-facing interface built upon $E$, rather than the main methodological focus of this work.

\begin{figure}
    \centering
    \includegraphics[width=1\linewidth]{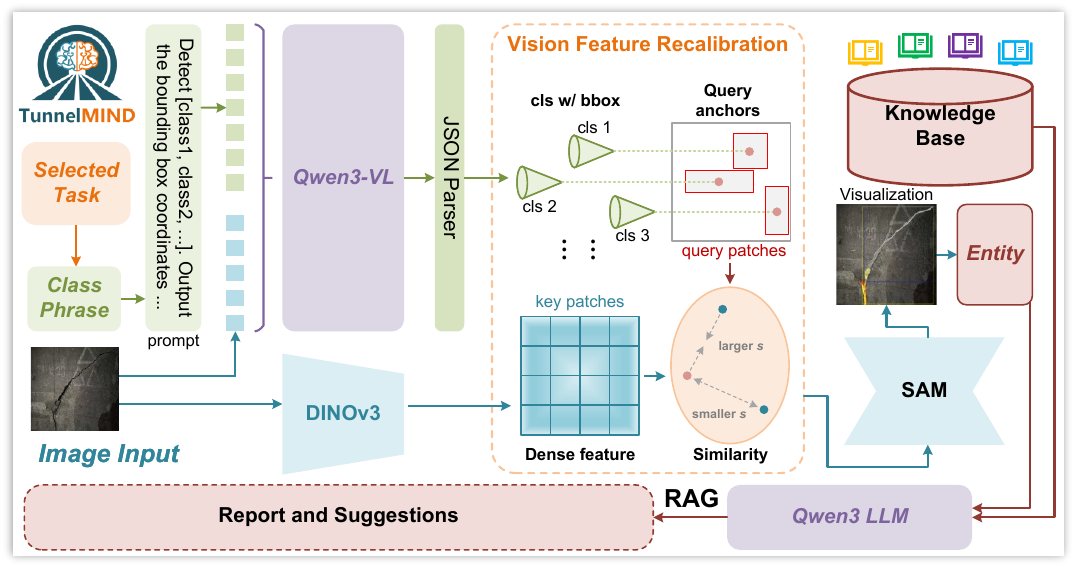}
    \caption{\textbf{Overview of TunnelMIND.} Qwen3-VL generates language-guided semantic proposals from the input image, selected task, and class phrases, while DINOv3 provides dense visual features for cross-model recalibration. The recalibrated prompts are converted by SAM into pixel-level masks, which are further organized into structured engineering entities and linked to retrieval-grounded explanation and report generation.}
    \label{fig:2}
\end{figure}

As illustrated in Figure~\ref{fig:2}, the pipeline begins with a vision-language model (Qwen3-VL \citet{bai2025qwen3}) that receives the selected task and class phrases as natural-language conditions and returns structured semantic proposals. These proposals provide an initial answer to what may be present, but they often remain spatially coarse and sensitive to hard negatives such as joints, stains, shadows, and repetitive textures. TunnelMIND therefore introduces an inference-time recalibration step that transfers these semantic anchors into the dense feature space of DINOv3, where local visual consistency is used to refine, expand, and regularize their spatial support. The recalibrated prompts are then fed into SAM to obtain masks with clearer boundaries and better geometric interpretability. Finally, the mask-level outputs are converted into structured engineering entities and linked to retrieval-grounded explanation and reporting. In this way, TunnelMIND is intended to establish a practical path from training-free perception to measurable and traceable tunnel inspection outputs, rather than to maximize task-specific benchmark accuracy through additional optimization.

\subsection{Language-guided semantic anchoring}\label{sec3.2}
The first stage of TunnelMIND is to generate class-aware candidate instances from natural-language task descriptions. Unlike conventional detectors trained for a fixed label space, the vision-language model in our framework is used as a language-guided semantic anchor generator rather than an end-to-end detector. This distinction is important because its role is to provide semantically plausible, structurally parsable, and task-reusable candidate regions, which are then refined by the subsequent visual recalibration module. Such a design is better suited to tunnel engineering scenarios, where task definitions may vary across applications, while the downstream system still requires a unified and verifiable interface.

Given image $I$ and category set $C=\{c_{1}, c_{2}, \ldots, c_{|C|}\}$, the VLM outputs a structured set of class-conditional bounding-box proposals:
\begin{equation}
    B_{c}=\{b_{c, i}\}_{i=1}^{N_{c}}, \qquad b_{c, i}=\left(x_{c, i}^{(1)}, y_{c, i}^{(1)}, x_{c, i}^{(2)}, y_{c, i}^{(2)}\right)
\end{equation}
where $N_{c}$ denotes the number of candidate instances for category $c$, and each $b_{c, i}$ represents the top-left and bottom-right coordinates of the corresponding bounding box. The overall proposal set $B$ is given by
\begin{equation}
    B=\bigcup_{c \in C} B_{c}
\end{equation}
To support stable multi-task reuse, all tasks share the same output schema, while only the task description and class list are changed. This allows rock debris segmentation, visible defect detection, GPR defect localization, road-damage detection, PPE inspection, and worker action monitoring to be represented under a unified interface without introducing task-specific heads.

\begin{figure}
    \centering
    \includegraphics[width=0.4\linewidth]{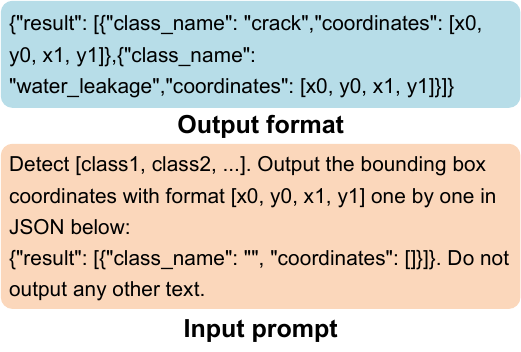}
    \caption{\textbf{Prompt and output format for language-guided semantic anchoring.} The model is required to return detected instances in a fixed JSON schema with class names and bounding box coordinates.}
    \label{fig:3}
\end{figure}

To improve output validity, we impose a constrained prompting and parsing strategy, as illustrated in Figure~\ref{fig:3}. The VLM is required to return its predictions in a fixed JSON-like structure containing an instance list, class names, and bounding-box coordinates. Each returned class name must belong to a predefined label set or its alias list, and each coordinate tuple must satisfy basic geometric validity constraints. Outputs that violate the schema are corrected only when the inconsistency is trivial, such as clipping coordinates to image boundaries, and are otherwise discarded when the error cannot be safely resolved. If the returned content is unparsable or misses the required \texttt{result} field, the system performs a single retry with synonym substitution for the queried class phrase. This retry is intended to recover semantically equivalent outputs while preventing uncontrolled prompt drift. In this way, the VLM interface remains robust enough for engineering deployment without relying on task-specific prompt engineering.

The purpose of this stage is to provide semantic seeds for subsequent spatial refinement. In tunnel imagery, open-vocabulary proposals are often sufficient to indicate potentially relevant regions, but they remain vulnerable to hard negatives and may exhibit inaccurate boundaries, duplicate candidates, or missing weak instances. For this reason, the semantic anchoring stage intentionally makes only a lightweight geometric commitment. It prioritizes semantic coverage and structural validity, while leaving spatial disambiguation and mask-quality improvement to the recalibration and segmentation stages that follow.

\subsection{Inference-time cross-model visual recalibration}\label{sec3.3}
A key design question is why TunnelMIND does not rely on a single foundation model for the entire pipeline. In tunnel inspection, this is mainly because semantic plausibility, spatial reliability, and geometric realization are not equally handled by the same component. A vision-language model is effective at answering what may be present, but its spatial support is often unstable under tunnel-specific hard negatives and weak boundaries. SAM, in contrast, can generate high-quality masks once reliable prompts are available, but it does not determine which regions are semantically relevant by itself. DINOv3 is introduced here as a dense visual support space because it provides locally consistent appearance representations that are well suited to anchor-centered retrieval and prompt refinement. From this perspective, the role of cross-model recalibration is exploit their complementary strengths: semantic anchoring from the VLM, dense visual support from DINOv3, and promptable geometric realization from SAM.

Although large vision-language models can provide semantically meaningful open-vocabulary proposals, their outputs are often insufficient for engineering use in tunnel environments. In practice, a VLM may detect only the most salient instances while missing weak, low-contrast, or visually fragmented targets. Even when the semantic category is roughly correct, the predicted regions may still suffer from inaccurate localization, boundary drift, and strong sensitivity to tunnel-specific hard negatives such as joints, stains, shadows, and repetitive textures. These limitations make the raw proposals unreliable for subsequent geometric quantification. By contrast, DINOv3 provides dense visual features with strong local consistency, where regions sharing similar texture or structural appearance tend to cluster in feature space even without explicit semantic supervision. This motivates a cross-model recalibration strategy: the VLM provides class-aware semantic anchors, and DINOv3 provides visually consistent spatial evidence for refining them.

\begin{figure}
    \centering
    \includegraphics[width=0.8\linewidth]{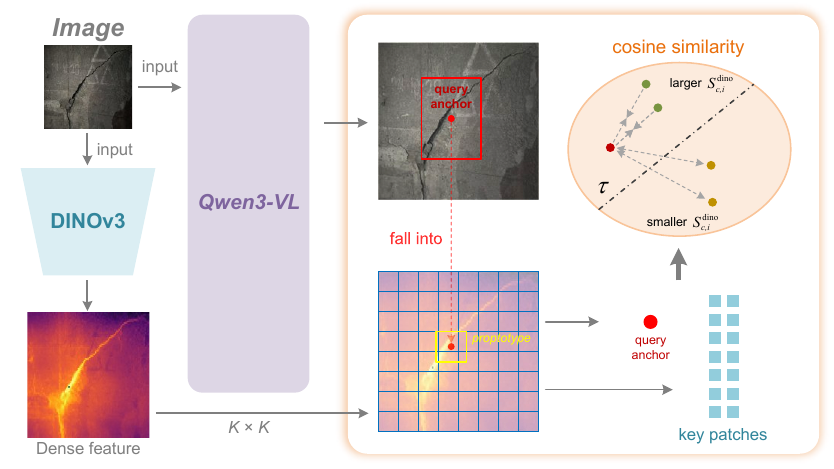}
    \caption{\textbf{Cross-model visual recalibration.} Qwen3-VL provides an initial query anchor, which is projected onto the DINOv3 dense feature map. The anchor-centered feature is then used to retrieve visually similar key patches by cosine similarity, forming a recalibrated spatial support for subsequent prompt generation.}
    \label{fig:4}
\end{figure}

\subsubsection{Anchor projection and dense support space}
Let $B_{c}=\{b_{c, i}\}_{i=1}^{N_{c}}$ denote the class-specific proposal set generated by the VLM for category $c$, where $b_{c, i}$ is the $i$-th bounding box. For each proposal, we define its center as the query anchor $a_{c, i}$:
\begin{equation}
    a_{c, i}=(a_{c,i}^{x},a_{c,i}^{y})=\left(\frac{x_{c, i}^{(1)}+x_{c, i}^{(2)}}{2}, \frac{y_{c, i}^{(1)}+y_{c, i}^{(2)}}{2}\right)
\end{equation}
The image $I$ is then fed into DINOv3 to obtain a dense feature map $D(I)$:
\begin{equation}
    D(I)=\{d_{u,v} \in \mathbb{R}^{D} \mid (u,v) \in \Omega\}
\end{equation}
where $\Omega$ denotes the spatial index set of the dense feature map, and $d_{u,v}\in\mathbb{R}^{D}$ is the $D$-dimensional local feature at position $(u,v)$.

Instead of retrieving directly on the native token grid of DINOv3, which varies across backbones, we introduce an explicit spatial partition with a controllable resolution $K$. Specifically, the image plane is uniformly divided into $K \times K$ key patches. The center of the patch $c_{m,n}$ at row $m$ and column $n$ is defined as
\begin{equation}
    c_{m,n}=\left(\frac{(n-1/2)W}{K}, \frac{(m-1/2)H}{K}\right)
\end{equation}
where $W$ and $H$ are the image width and height, respectively. For each patch $p_{m,n}$, we average-pool the dense DINOv3 features within its covered region to obtain a patch-level representation $k_{m,n}$. This yields a unified dense retrieval space with task-independent and backbone-independent spatial resolution. For a given query anchor $a_{c,i}$, we first locate the grid cell that contains it:
\begin{equation}
    n^*=\mathrm{clip}\left(\left\lfloor \frac{a_{c,i}^{x}}{W}K \right\rfloor + 1,1,K\right), \qquad
    m^*=\mathrm{clip}\left(\left\lfloor \frac{a_{c,i}^{y}}{H}K \right\rfloor + 1,1,K\right)
\end{equation}
where $\lfloor \cdot \rfloor$ denotes the floor operation and $\mathrm{clip}(\cdot,1,K)$ restricts the index to the valid grid range. In this way, an initially semantic query is projected into a dense, task-independent visual support space.

\subsubsection{Local prototype construction and support retrieval}
The corresponding patch feature $k_{m^*,n^*}$ can be directly used as the query representation. However, because the VLM anchor may exhibit small spatial bias, especially for elongated or weak defects such as cracks, directly retrieving with a single patch feature can lead to support drift. To construct a more stable anchor-centered prototype, we aggregate the features within an $r \times r$ neighborhood centered at $(m^*,n^*)$:
\begin{equation}
    q_{c,i}=\mathrm{Avg}\{k_{m,n} \mid (m,n) \in \mathcal{N}(m^*,n^*)\}
\end{equation}
where $\mathcal{N}(m^*,n^*)$ denotes the local neighborhood. This local averaging suppresses feature noise and reduces retrieval drift caused by imperfect anchor placement. Importantly, this prototype is constructed on the fly during inference from the anchor neighborhood itself, rather than learned through any task-specific optimization.

We then compute the cosine similarity between the query representation $q_{c,i}$ and all key patches:
\begin{equation}
    s_{c,i}^{\mathrm{dino}}(m,n)=\cos\left(q_{c,i},k_{m,n}\right)
\end{equation}
A preliminary support set is first obtained by similarity thresholding:
\begin{equation}
    \widehat{P}_{c,i}=\{(m,n) \mid s_{c,i}^{\mathrm{dino}}(m,n) \ge \tau\}
\end{equation}
where $\tau$ is a similarity threshold. Since thresholding alone may still retain too many positive regions in cluttered tunnel backgrounds, we further rank the retained patches in $\widehat{P}_{c,i}$ by similarity score and keep at most the $\mathrm{Top}$-$M$ elements. The final positive support set is therefore defined as
\begin{equation}
    P_{c,i}=\mathrm{Top}\text{-}M\left(\widehat{P}_{c,i}\right)
\end{equation}
where $\mathrm{Top}$-$M$ returns up to $M$ highest-scoring patches after threshold filtering. For each retained $(m,n) \in P_{c,i}$, its center $c_{m,n}$ is used as a positive prompt point, forming
\begin{equation}
    \Pi^{+}_{c,i}=\{c_{m,n} \mid (m,n) \in P_{c,i}\}
\end{equation}
In this way, the semantic cue provided by the VLM is transferred into a visually coherent support region in the DINOv3 feature space.

\subsubsection{Prompt regularization for segmentation}
To further improve prompt quality for SAM, we consider two prompting modes. In the positive-only mode, all points in $\Pi^{+}_{c,i}$ are treated as foreground prompts, which is suitable when the target boundary is relatively clear and the surrounding background is simple. In the positive-negative mode, an additional negative prompt set $\Pi^{-}_{c,i}$ is introduced by sampling from the lowest-similarity patches or from regions outside the proposal neighborhood. These negative points suppress background leakage and reduce over-segmentation, particularly for defect categories with fuzzy appearance or strong texture interference. When multiple recalibrated proposals belong to the same category, non-maximum suppression (NMS) is finally applied to reduce duplicates before the segmentation stage \citet{hosang2017learning}.

The recalibration module combines two complementary sources of information: the VLM estimates what is likely to be present, while DINOv3 identifies where visually consistent evidence is located. As a result, raw open-vocabulary proposals are transformed into more reliable prompts for segmentation and downstream engineering quantification.

\subsection{Entity-centric engineering reconstruction}\label{sec3.4}
\subsubsection{Prompted geometric realization}
\begin{figure}
    \centering
    \includegraphics[width=0.55\linewidth]{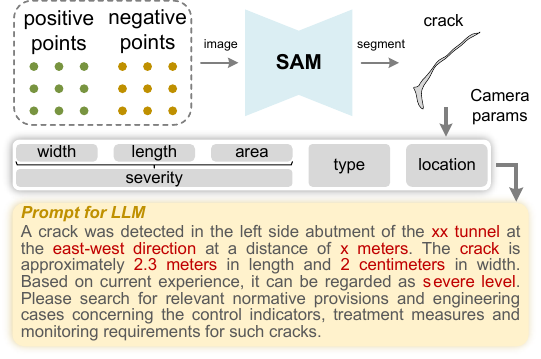}
    \caption{\textbf{Engineering entity construction and retrieval-grounded reporting interface.} Recalibrated positive and negative prompts are fed into SAM to obtain target masks, from which category, structural location, geometric attributes, and severity are derived. These structured entities are then converted into retrieval queries for evidence-grounded explanation and engineering-readable report generation.}
    \label{fig:5}
\end{figure}

The recalibrated prompt sets produced in Section~\ref{sec3.3} are next used to obtain pixel-level target delineation. Let $\Pi^{+}_{c,i}$ and $\Pi^{-}_{c,i}$ denote the positive and negative prompt sets associated with the $i$-th recalibrated proposal of category $c$ (Figure~\ref{fig:5}). These prompts, optionally combined with the corresponding proposal box, are fed into SAM to generate a binary mask:
\begin{equation}
    m_{c,i}=\Psi_{\mathrm{sam}}\left(I, \Pi^{+}_{c,i}, \Pi^{-}_{c,i}, \tilde{b}_{c,i}\right)
\end{equation}
where $\tilde{b}_{c,i}$ denotes the recalibrated spatial support inherited from the proposal refinement stage. Compared with directly segmenting from raw VLM boxes, the recalibrated prompts provide denser and more spatially coherent foreground evidence, which is particularly beneficial in tunnel imagery where weak boundaries and background leakage are common. In this sense, SAM is not used as an isolated segmenter, but as a geometric realization module that converts visually recalibrated cues into engineering-interpretable instance masks.

\subsubsection{Task-aware geometric measurement}
Once the mask $m_{c,i}$ is obtained, TunnelMIND extracts task-dependent geometric attributes for downstream engineering interpretation. For crack-like defects, we compute the mask skeleton and use it to estimate the effective crack length $L_i$. The crack width $W_i$ is then estimated from the distribution of normal distances measured along the skeleton, and a representative width statistic is retained for subsequent grading. For area-like defects such as seepage, spalling, or road damage, we compute the mask area $A_i$, together with auxiliary shape descriptors such as maximum diameter and compactness. These geometric quantities provide a more engineering-relevant representation than raw pixel masks, since downstream maintenance decisions are typically made based on measurable physical indicators rather than on segmentation contours alone.

\subsubsection{Structured entity reconstruction and severity grounding}
Based on the semantic category and the extracted geometric measurements, each instance is converted into an engineering entity:
\begin{equation}
    e_i=[t_i,l_i,g_i,s_i,c_i]
\end{equation}
where $t_i$ denotes the engineering type, $l_i$ denotes the structural location, $g_i$ denotes the geometric attribute set, $s_i$ denotes the severity level, and $c_i$ denotes the contextual information. Specifically, the engineering type $t_i$ is obtained by mapping the open-vocabulary class output of the VLM to a normalized engineering taxonomy, such as crack, seepage, spalling, pothole, or other task-dependent categories. The location term $l_i$ is determined by combining the proposal box or mask centroid with pre-calibrated camera geometry, allowing the instance to be expressed in tunnel-oriented structural coordinates such as left wall, right wall, crown, or road surface. The geometric descriptor $g_i$ contains quantities such as $L_i$, $W_i$, and $A_i$, depending on the defect category. This entity-centric representation explicitly bridges the gap between vision outputs and the forms of information required in engineering records and maintenance workflows.

The severity label $s_i$ is assigned by comparing the extracted geometric indicators with predefined thresholds derived from engineering specifications or expert criteria. Rather than asking an unconstrained language model to infer whether a defect is minor or severe, TunnelMIND first grounds severity in measurable attributes and only then passes the structured result to the downstream reporting interface. This design has two advantages. First, it preserves traceability between visual evidence and engineering judgment. Second, it prevents multiple sources of uncertainty from being collapsed into a single free-form textual conclusion. In addition, the context term $c_i$ records auxiliary information such as recalibration confidence, construction section or tunnel section ID, and acquisition time, so that each entity remains connected to its source evidence and deployment setting. From a methodological perspective, the role of this stage is to transform segmentation masks into structured engineering instances that are quantitative, localized, and compatible with downstream retrieval and report assistance.

\subsection{Deployment-facing retrieval and report interface}\label{sec3.5}
The engineering entities constructed in Section~\ref{sec3.4} are finally mapped to retrieval-grounded explanation and inspection assistance. Rather than asking a large language model to infer maintenance actions directly from visual outputs, TunnelMIND introduces a constrained generation layer in which the LLM operates primarily as a synthesis and reporting engine conditioned on retrieved expert knowledge. This design is motivated by a practical consideration in tunnel inspection: defect cases are heterogeneous, annotated decision data are scarce, and engineering suggestions must remain aligned with standards, technical guidelines, and prior cases. Therefore, the purpose of this stage is not to automate engineering decision-making, but to connect structured visual evidence with traceable textual knowledge and generate engineering-readable explanations and advisory suggestions under explicit constraints.

Let $K=\{d_k\}_{k=1}^{N}$ denote the expert knowledge base, where each document fragment $d_k$ is obtained by structured splitting of engineering materials such as national codes, industry standards, design manuals, defect investigation reports, strengthening specifications, and accident analysis documents. Each fragment is encoded by a text embedding model $E_{\text{text}}$ into a vector
\begin{equation}
    z_k=E_{\text{text}}(d_k)
\end{equation}
and stored in a vector database together with metadata such as defect category, applicable structural part, and document source. This organization allows the knowledge base to support defect-aware retrieval instead of generic semantic search. For each engineering entity $e_i$, TunnelMIND converts its structured attributes into a natural-language retrieval query $q_i$ using a predefined template. The query summarizes key information such as defect type, structural location, geometric measurements, and preliminary severity level, and may additionally include project identifiers such as tunnel section or mileage when available. The query is then embedded as
\begin{equation}
    z_{q_i}=E_{\text{text}}(q_i)
\end{equation}
and the top-$k$ most similar fragments $\{d_{i1}, d_{i2}, \ldots, d_{ik}\}$ are retrieved from the vector database by nearest-neighbor search. By retrieving standards and prior cases that match the measured entity description, the system grounds downstream interpretation in external evidence rather than in free-form language priors.

Given the structured entity $e_i$, the retrieved evidence set $\{d_{ij}\}_{j=1}^{k}$, and a task instruction, TunnelMIND invokes an LLM to generate two types of outputs: an inspection-oriented explanation and an advisory engineering suggestion. To improve safety and traceability, we impose three generation constraints. First, the LLM is encouraged to explicitly refer to the retrieved source fragments whenever possible, so that the generated explanation remains linked to identifiable engineering evidence. Second, when the measured geometric attributes are close to grading thresholds or when the retrieved evidence is insufficiently specific, the model is required to state the uncertainty explicitly and recommend on-site reinspection instead of forcing a categorical conclusion. Third, the generated suggestions are phrased as advisory options rather than mandatory commands, with the final decision reserved for qualified engineers. If no sufficiently matched evidence is found in the knowledge base, the system outputs a conservative fallback statement indicating that no clear recommendation is available from the current knowledge base and that case-specific judgment by professional engineers is required. Through these constraints, this stage should be understood as a deployment-facing reporting interface rather than an autonomous decision module. Its role is to enable reliable perception results to enter engineering workflows in an explainable, traceable, and practically useful form. \textit{Detailed knowledge-base composition, chunk construction, retrieval settings, and query templates are provided in Supplementary C.}

\section{Experiments}\label{sec4}
This section evaluates TunnelMIND from three perspectives: inspection reliability, engineering entity quality, and retrieval-grounded reporting utility. Since the proposed framework is centered on cross-model visual recalibration rather than task-specific supervised optimization, the experiments are organized around mechanism validation instead of benchmark enumeration. We first report the main results on representative tunnel inspection tasks, then examine robustness under hard-negative conditions and cross-task interface transferability, and finally evaluate whether the entity-centric representation and retrieval-grounded reporting layer improve downstream engineering usefulness.

\subsection{Experimental settings}
To evaluate TunnelMIND under realistic tunnel inspection conditions, we design the experiments from three complementary perspectives: representative-task performance, cross-task interface transferability, and engineering-oriented downstream utility. Rather than treating all six tasks as equally weighted benchmarks, we select several representative tasks for primary evaluation and use the remaining tasks to assess interface-level transferability. This design is consistent with the goal of this work, which is not to build a separate state-of-the-art model for each task, but to examine whether the proposed training-free pipeline can reliably transform open-vocabulary proposals into engineering-ready entities across heterogeneous tunnel inspection settings. In addition, because tunnel imagery is frequently affected by shadows, joints, stains, repetitive textures, and low-contrast targets, we further construct a hard-negative subset to evaluate robustness under interference-heavy conditions.

\begin{figure}
    \centering
    \includegraphics[width=0.9\linewidth]{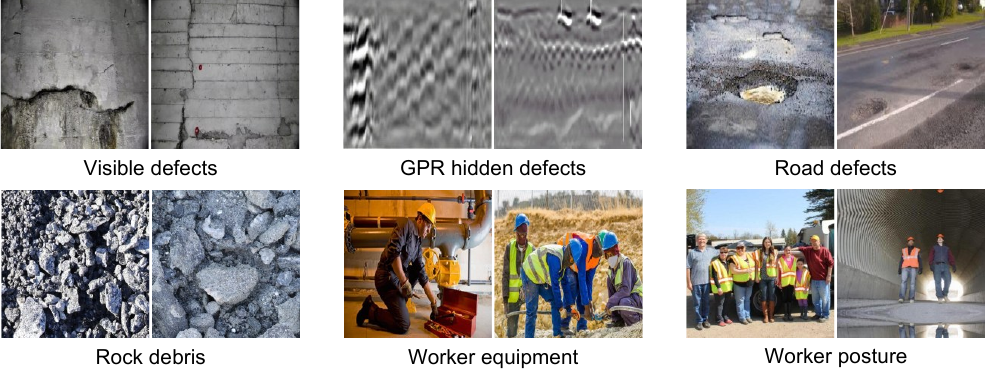}
    \caption{\textbf{Example images from the six tasks considered in this work}, including visible defects, GPR hidden defects, road defects, worker pose, rock slag, and PPE inspection. These samples illustrate the diversity of imaging modality, target appearance, and scene complexity across tunnel-related construction and operation scenarios.}
    \label{fig:6}
\end{figure}

\textbf{Datasets and tasks.} We consider six tunnel-related tasks spanning both construction and operation scenarios, including rock slag segmentation (Rock), visible lining defect detection (Visible), GPR hidden defect detection (GPR), road defect detection (Road), PPE inspection (PPE), and worker pose recognition (Pose). Among them, visible defects, GPR defects, and road defects are selected as representative tasks for the main analysis, since they jointly reflect semantic ambiguity, texture interference, irregular geometry, and engineering quantification requirements. The remaining tasks are used to assess cross-task transferability. Example images for all six tasks are shown in Figure~\ref{fig:6}, and the corresponding sample allocation is summarized in Table~\ref{tab:1}.

\begin{table}
    \centering
    \setlength{\heavyrulewidth}{1.2pt}
    \setlength{\lightrulewidth}{0.6pt}
    \caption{\textbf{Illustrative dataset and evaluation protocol summary.} The PPE inspection utilized the publicly available construction ppe dataset. Worker pose data was sampled from MS COCO \citet{lin2014microsoft}, while other task-related data were collected independently.}
    \begin{tabular}{lllccc}
    \toprule
    \textbf{Task} & \textbf{Scenario} & \textbf{Annotation} & \textbf{Train images} & \textbf{Val images} & \textbf{Main Metric} \\
    \midrule
    Rock debris & Construction & mask & 270 & 68 & IoU \\
    Visible defects & Operation & bbox, mask & 715 & 179 & P / R / F1 \\
    GPR hidden defects & Operation & bbox, mask & 930 & 233 & P / R / F1 \\
    Road defects & Both & bbox, mask & 4823 & 401 & P / R / F1 \\
    PPE inspection & Both & bbox & 1132 & 143 & P / R / F1 \\
    Worker pose & Both & keypoints & 2592 & 648 & P / R / F1 \\
    \bottomrule
    \end{tabular}
    \label{tab:1}
\end{table}

To better reflect engineering deployment settings, the data are split by tunnel section or acquisition source whenever possible, rather than by purely random image-level sampling. This design reduces the risk of overestimating generalization due to highly similar neighboring frames. The hard-negative subset is constructed from validation samples with strong shadows, joints, stains, reflections, or repetitive structural textures.

\textbf{Evaluation metrics.} For perception performance, we report intersection over union (IoU) for rock slag segmentation and precision (P), recall (R), and F1 score (F1) for the representative detection tasks. For entity-level engineering evaluation, we report geometric measurement errors, structural location accuracy, severity classification performance, and entity completeness. For report assistance, we evaluate retrieval quality together with LLM-as-judge report quality, including usefulness, safety, and clarity.

\textbf{Engineering ground-truth protocol.} Since the purpose of TunnelMIND is not only image-level localization but also engineering-ready interpretation, we designed the entity-level evaluation under an explicit task-level engineering protocol. Physical measurements were derived from controlled acquisition settings with calibrated pixel-to-physical mapping, so that the predicted geometric attributes could be converted into measurable quantities under a unified task-specific scale. For visible lining defects and road defects, the imaging configuration was fixed within each task, allowing crack length, representative width, and damaged area to be evaluated in physical units rather than only in pixels. Structural location labels were assigned by combining acquisition logs with predefined imaging geometry, so that each detected entity could be mapped to engineering regions such as wall-side, crown, or road surface zone. Severity labels were then assigned by rule-based comparison against predefined engineering handbooks and maintenance criteria using the measured geometric attributes. Under this protocol, the entity-level evaluation was intended to examine whether the predicted outputs could support measurable, localizable, and gradeable engineering interpretation, rather than only whether they overlapped with image annotations. \textit{Detailed calibration formulas, location assignment rules, severity criteria, and annotation verification procedures are provided in Supplementary A.} To avoid overstating image-level performance as engineering usability, all entity-level metrics in this work should be interpreted under this protocolized acquisition-and-calibration setting.

\textbf{Report-quality scoring protocol.} Since large-scale formal engineering review was impractical in a fully controlled setting, we adopted a model-based judging protocol for comparative evaluation. Three large language models, namely DeepSeek \citet{liu2024deepseek}, Qwen3, and Kimi \citet{team2025kimi}, were provided with the same domain knowledge materials and a unified scoring rubric covering usefulness, safety, and clarity, and their averaged scores were taken as preliminary report scores. To reduce occasional scoring instability, all report cases were further manually verified by two co-authors according to the same rubric. Manual correction was applied only when a score was clearly inconsistent with the report content under the predefined rubric; otherwise, the original averaged score was retained. When the two verifiers did not reach agreement, the original averaged LLM score was kept. This protocol was intended to provide a stable comparative signal across methods rather than to replace formal engineering assessment.

\textbf{Visual prompting protocol for GPR.} For GPR hidden defect detection, Qwen3-VL-4B-Instruct and Rex-Omni \citet{jiang2025detect} are evaluated using visual prompting rather than purely text-driven querying, because GPR defect patterns are highly abstract and often lack stable correspondence to explicit natural-language semantics. Specifically, one fixed reference example is assigned to each defect category and used throughout the evaluation as the visual prompt. Given the reference example, the model is instructed to localize visually similar abnormal responses in the target radar image. To ensure fairness, the same reference example, prompt template, and output parsing rule are shared across the compared multimodal baselines within the same category. The returned responses are converted into coarse spatial cues, such as boxes or point-level hints depending on the model output format, and are further paired with SAM when required by the baseline configuration. Under this protocol, the comparison focuses on whether multimodal large models can support training-free similarity-based localization in weak-semantic radar imagery, rather than conventional language-driven open-vocabulary detection. \textit{The detailed GPR visual prompting setup, including reference-target concatenation, parsing rules, and the fairness constraints across compared baselines, is provided in Supplementary B.}

\textbf{Implementation details.} All training-free methods use frozen foundation models without task-specific fine-tuning. Unless otherwise stated, the same prompting template and shared recalibration configuration are used across tasks. The VLM is instantiated using Qwen3-VL-4B-Instruct. For visual feature extraction, we adopt DINOv3 with a ViT-S+/16 \citet{vaswani2017attention} backbone. Segmentation is performed using SAM3 \citet{carion2025sam}. All models are deployed on 2 NVIDIA Geforce RTX 3090 GPUs. The input image resolution is constrained such that the longer side does not exceed 768 pixels. The number of sampled key patches is set to $K=24$, and the neighborhood radius is defined as $r=5$. The similarity threshold is fixed at $\tau = 0.6$, and the maximum number of retained positive prompts is set to $\mathrm{Top}\text{-}M=5$. The number of negative prompts is also capped at $5$. Category-wise non-maximum suppression with an IoU threshold of $0.5$ is finally applied before segmentation or entity construction. \textit{Additional implementation and deployment details, including software environment and stage-wise dual-GPU configuration, are provided in Supplementary D.}

\subsection{Compared methods}
To comprehensively evaluate TunnelMIND under both performance-oriented and deployment-oriented settings, we compare it with two groups of methods: supervised task-specific models and training-free foundation-model-based methods. The former serve as strong performance references under fully supervised optimization, while the latter are used to examine the practical effectiveness of different training-free paradigms without task-specific parameter updates.

For supervised methods, we select DEIMv2 \citet{huang2025real}, RT-DETRv4 \citet{liao2025rt}, YOLOv11 \citet{khanam2024yolov11}, and YOLOv12 \citet{tian2025yolov12} as representative baselines. The YOLO family provides strong engineering practicality and deployment efficiency, whereas RT-DETRv4 and DEIMv2 represent recent high-performance end-to-end detection frameworks. All supervised baselines are trained on the corresponding task annotations and are used as upper-bound references for task-specific optimization.

For training-free methods, we compare against GroundingDINO \citet{liu2024grounding}, Qwen3-VL-4B-Instruct, and Rex-Omni. GroundingDINO represents a typical open-vocabulary detector that produces candidate boxes directly from text prompts, while Qwen3-VL-4B-Instruct and Rex-Omni represent multimodal large-model based inspection systems with stronger visual grounding capability. Since these methods mainly provide coarse boxes or point-level cues, we further pair them with SAM through box-based or point-based prompting when applicable, so as to evaluate whether their outputs can be effectively transformed into usable segmentation results in tunnel scenes.

\subsection{Main results on representative defect-oriented tasks}\label{sec4:main}
Table~\ref{tab:2} reports the results on three representative tasks, namely visible defect detection, GPR hidden defect detection, and road defect detection. Overall, supervised methods still achieve the best performance, with RT-DETRv4-X showing the strongest overall results across the three tasks. This indicates that task-specific models remain advantageous when annotations are sufficient and task boundaries are clearly defined. In contrast, training-free methods yield lower absolute performance, although clear differences can still be observed among them.

\begin{table}
\centering
\setlength{\heavyrulewidth}{1.2pt}
\setlength{\lightrulewidth}{0.6pt}
\caption{\textbf{Main results on representative defect-oriented tunnel inspection tasks.} For the GPR task, the results of Qwen3-VL-4B and Rex-Omni are obtained under the visual prompting protocol, where one fixed reference example is used for each defect category to support similarity-based localization in weak-semantic radar imagery. GroundingDINO is evaluated under text-driven open-vocabulary detection. The same fixed reference example, prompt template, and parsing rule are used across the compared multimodal baselines within each GPR category. Bold and Underline indicate the best and second-best results, respectively.}
\begin{tabular}{llccccccccc}
\toprule
\multirow{2}{*}{\textbf{Type}} & \multirow{2}{*}{\textbf{Method}} 
& \multicolumn{3}{c}{\textbf{Visible}} 
& \multicolumn{3}{c}{\textbf{GPR}} 
& \multicolumn{3}{c}{\textbf{Road}} \\
\cmidrule(lr){3-5} \cmidrule(lr){6-8} \cmidrule(lr){9-11}
& 
& \textbf{P} & \textbf{R} & \textbf{F1} 
& \textbf{P} & \textbf{R} & \textbf{F1} 
& \textbf{P} & \textbf{R} & \textbf{F1} \\
\midrule
\multirow{2}{*}{Training-based}
& YOLOv12-X   & 0.84 & 0.82 & 0.83 & 0.91 & 0.87 & 0.89 & 0.88 & 0.78 & 0.83 \\
& RT-DETRv4-X & 0.87 & 0.83 & 0.85 & 0.92 & 0.87 & 0.89 & 0.89 & 0.79 & 0.84 \\
\midrule
\multirow{4}{*}{Training-free}
& GroundingDINO & 0.60 & 0.53 & 0.56 & 0.12 & 0.09 & 0.10 & 0.63 & 0.51 & 0.56 \\
& Rex-Omni      & \underline{0.63} & \underline{0.59} & \underline{0.61} & \underline{0.74} & 0.72 & \underline{0.73} & 0.65 & 0.52 & 0.58 \\
& Qwen3-VL-4B   & 0.61 & 0.54 & 0.57 & 0.65 & \underline{0.74} & 0.69 & \underline{0.69} & \underline{0.60} & \underline{0.64} \\
& \textbf{TunnelMIND} & \textbf{0.68} & \textbf{0.67} & \textbf{0.68} & \textbf{0.76} & \textbf{0.81} & \textbf{0.78} & \textbf{0.76} & \textbf{0.69} & \textbf{0.72} \\
\bottomrule
\end{tabular}
\label{tab:2}
\end{table}

\begin{figure}
    \centering
    \includegraphics[width=1\linewidth]{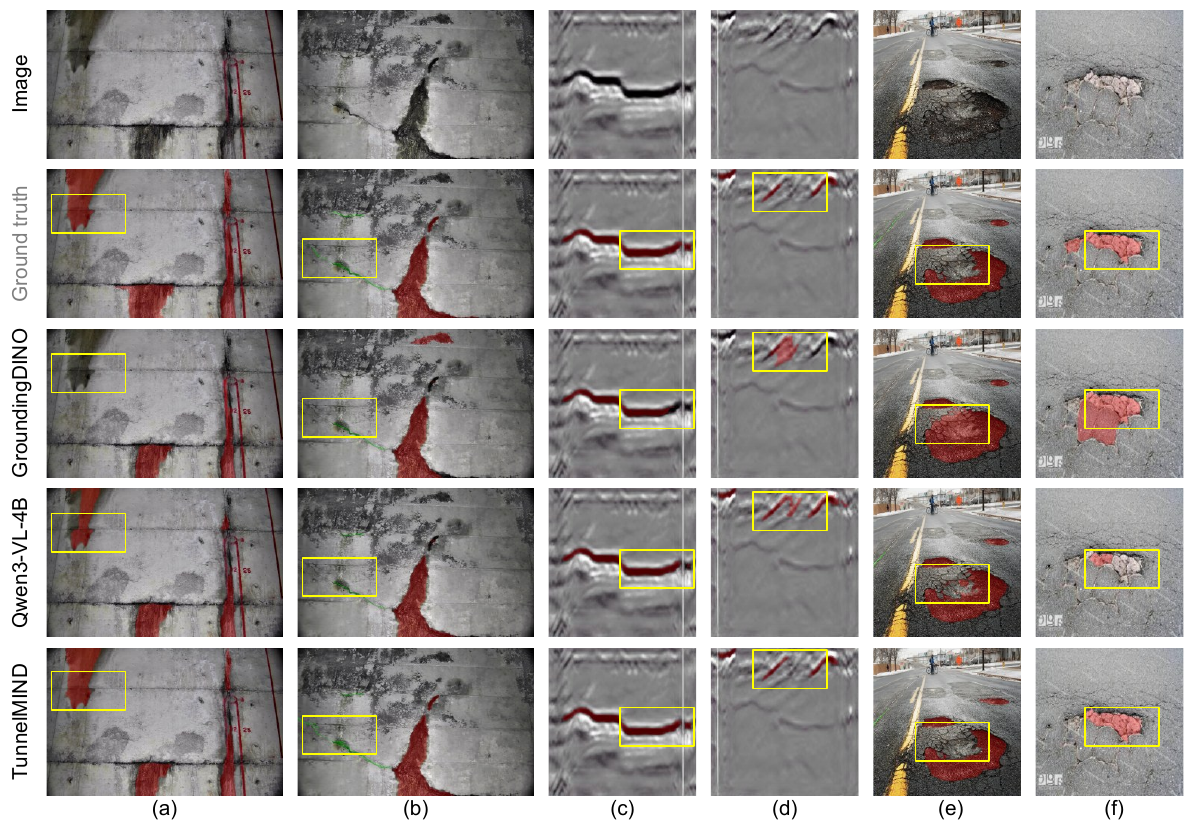}
    \caption{\textbf{Qualitative comparison on representative tasks.} (a)(b) show visible lining defects, (c)(d) show GPR hidden defect cases under the visual prompting protocol, and (e)(f) show road defect cases. From top to bottom are the input image, ground truth, GroundingDINO, Qwen3-VL-4B, and TunnelMIND. Compared with the training-free baselines, TunnelMIND produces spatially more complete and visually more coherent defect regions, especially for elongated cracks, continuous GPR abnormal responses, and irregular road-damage areas, indicating that cross-model visual recalibration improves the usability of training-free inspection outputs.}
    \label{fig:7}
\end{figure}

Among the training-free methods, GroundingDINO performs worst, especially on the GPR task, where both precision and recall are very low. This is expected because GPR defect patterns usually lack a stable correspondence to explicit natural-language semantics, making purely text-driven open-vocabulary detection poorly suited to radar imagery. For this reason, we evaluate Rex-Omni and Qwen3-VL-4B with visual prompting on GPR, where reference examples are used to retrieve visually similar abnormal responses. The results suggest that multimodal large models possess transferable visual representations that can support similarity-based localization in weak-semantic radar data. However, on visible and road defect tasks, these methods remain sensitive to complex backgrounds, weak boundaries, and hard negatives, leading to limited spatial stability and incomplete defect recovery. By contrast, TunnelMIND achieves the best performance among all training-free methods, reaching F1 scores of 0.68, 0.78, and 0.72 on Visible, GPR, and Road, respectively. In particular, its recall reaches 0.81 on GPR, indicating that cross-model visual recalibration can suppress false positives while recovering part of the weak missed targets. These results show that semantic proposals alone are insufficient for reliable training-free tunnel inspection, whereas combining language-guided candidates with dense visual consistency can substantially improve the usability of defect-oriented inspection outputs.

These quantitative results are further supported by the qualitative comparisons in Figure~\ref{fig:7}. As shown in Figure~\ref{fig:7}, (a)(b) correspond to Visible, (c)(d) correspond to GPR, and (e)(f) correspond to Road. GroundingDINO can respond to some salient targets in the Visible and Road tasks, but it is easily affected by background textures and noise, leading to spatial offsets and incomplete boundaries; its performance is even weaker on GPR imagery. In comparison, Qwen3-VL-4B with visual prompting can more stably retrieve abnormal regions similar to the reference examples, but it still suffers from missed weak targets and coarse defect recovery in complex scenes. TunnelMIND produces results that are more consistent with the ground truth, especially for elongated cracks, continuous GPR abnormal responses, and complete road-damage regions, demonstrating that cross-model visual recalibration improves both spatial localization and result completeness under training-free settings.

\subsection{Robustness under hard-negative conditions}
Since TunnelMIND is motivated by the limited spatial reliability of language-guided proposals under tunnel-specific hard negatives, robustness under interference-heavy conditions serves as a direct validation of the proposed recalibration mechanism. Table~\ref{tab:3} reports the results on the hard-negative subset, which contains strong shadows, joints, stains, reflections, or repetitive textures. All compared methods suffer noticeable degradation on these samples. Nevertheless, TunnelMIND still achieves the best average F1 of 0.66, outperforming Qwen3-VL-4B w/ SAM(point) at 0.61 and GroundingDINO w/ SAM(box) at 0.35. This indicates that the advantage of TunnelMIND lies not only in average-case performance, but also in improved robustness under complex engineering backgrounds.

\begin{table}
    \centering
    \setlength{\heavyrulewidth}{1.2pt}
    \setlength{\lightrulewidth}{0.6pt}
    \caption{\textbf{Results on the hard-negative subset.} The terms `box' and `point' denote box and point annotations used for SAM segmentation, respectively. `Avg.' indicates the average value per row.}
    \begin{tabular}{lcccc}
    \toprule
    \textbf{Method} & \textbf{Visible F1} & \textbf{GPR F1} & \textbf{Road F1} & \textbf{Avg.} \\
    \midrule
    GroundingDINO w/ SAM (box) & 0.49 & 0.04 & 0.52 & 0.35 \\
    Rex-Omni                   & 0.54 & 0.55 & 0.53 & 0.54 \\
    Qwen3-VL-4B w/ SAM (box)   & 0.57 & 0.59 & 0.61 & 0.59 \\
    Qwen3-VL-4B w/ SAM (point) & \underline{0.58} & \underline{0.61} & \underline{0.64} & \underline{0.61} \\
    \textbf{TunnelMIND}        & \textbf{0.64} & \textbf{0.66} & \textbf{0.68} & \textbf{0.66} \\
    \bottomrule
    \end{tabular}
    \label{tab:3}
\end{table}

\begin{figure}
    \centering
    \includegraphics[width=1\linewidth]{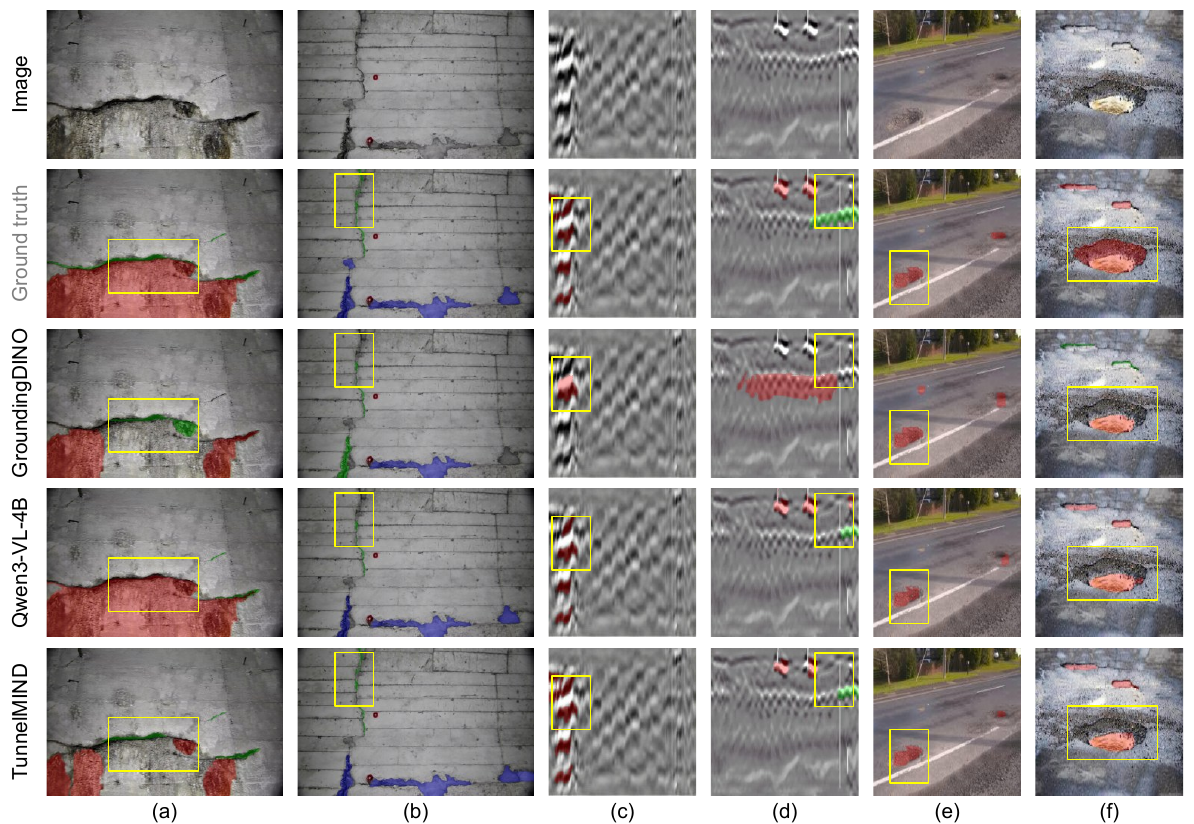}
    \caption{\textbf{Qualitative comparison on hard-negative samples.} (a)(b) show visible defect cases with strong texture interference, (c)(d) show GPR hidden defect cases with abnormal radar-wave interference, and (e)(f) show road defect cases with granular background or low resolution. From top to bottom are the input image, ground truth, GroundingDINO, Qwen3-VL-4B, and TunnelMIND. Under interference-heavy conditions, TunnelMIND suppresses background distraction more effectively and preserves more complete defect morphology, showing that the proposed recalibration mechanism improves the robustness of training-free inspection in complex tunnel scenes.}
    \label{fig:8}
\end{figure}

Figure~\ref{fig:8} provides additional hard-negative visualizations. Here, (a)(b) show Visible cases with strong texture interference, (c)(d) show GPR cases with abnormal radar-wave interference, and (e)(f) show Road cases with granular backgrounds or low resolution. Under these challenging conditions, GroundingDINO is prone to false positives, region offsets, and boundary distortion, while Qwen3-VL-4B can still respond to the target region but often includes background clutter or incomplete defect shapes. In contrast, TunnelMIND remains more focused on the true defect area and preserves more complete target morphology across all three types of hard-negative scenes, further confirming that the proposed visual recalibration mechanism can effectively suppress interference from complex backgrounds and abnormal responses.

\subsection{Cross-task interface transferability}
To examine the breadth of the proposed pipeline beyond the representative defect tasks, we further evaluate TunnelMIND on additional tunnel-related scenarios using the same training-free interface. The goal of this analysis is not to claim uniform superiority across all task types, but to test whether the proposed proposal-recalibration-entity pipeline can be reused across heterogeneous inspection settings with limited task-specific adaptation.

Table~\ref{tab:4} reports the results on six tunnel-related tasks, including rock slag segmentation, visible lining defect detection, GPR hidden defect localization, road defect detection, PPE inspection, and worker pose recognition. Overall, TunnelMIND maintains relatively stable outputs across multiple tasks under a unified training-free setting. In particular, it performs competitively on Rock, Visible, GPR, Road, and PPE, and is especially strong on the three representative defect-oriented tasks, namely Visible, GPR, and Road. These results indicate that the same semantic-proposal and visual-recalibration interface can be reused across heterogeneous tunnel inspection scenarios without task-specific retraining.

\begin{table*}
\centering
\setlength{\heavyrulewidth}{1.2pt}
\setlength{\lightrulewidth}{0.6pt}
\caption{Cross-task interface transferability under a unified training-free inspection setting. $\dagger$ denotes SAM combined with box prompt. $\ddagger$ indicates results obtained under the GPR visual prompting protocol. Bold and underline indicate the best and second-best results, respectively. N/A means that the compared method does not support the corresponding task type.}
\begin{tabular}{llcccccc}
\toprule
\textbf{Type} & \textbf{Method} & \textbf{Rock IoU} & \textbf{Visible P/R} & \textbf{GPR P/R} & \textbf{Road P/R} & \textbf{PPE P/R} & \textbf{Pose P/R} \\
\midrule
\multirow{4}{*}{Training-based}
& YOLOv11-X   & 0.79 & 0.82/0.79 & 0.89/0.85 & 0.85/0.74 & 0.93/0.91 & 0.82/0.76 \\
& YOLOv12-X   & 0.79 & 0.84/0.82 & 0.91/0.87 & 0.88/0.78 & 0.95/0.93 & 0.84/0.79 \\
& DEIMv2-X    & 0.80 & 0.85/0.82 & 0.90/0.86 & 0.88/0.77 & 0.96/0.93 & 0.87/0.82 \\
& RT-DETRv4-X & 0.83 & 0.87/0.83 & 0.92/0.87 & 0.89/0.79 & 0.96/0.93 & 0.88/0.82 \\
\midrule
\multirow{4}{*}{Training-free}
& GroundingDINO & 0.50$^\dagger$ & 0.60/0.53 & 0.12/0.09 & 0.63/0.51 & 0.83/0.76 & N/A \\
& Rex-Omni      & 0.63$^\dagger$ & \underline{0.63/0.59} & \underline{0.74}/0.72$^\ddagger$ & 0.65/0.52 & 0.86/0.79 & \underline{0.76/0.72} \\
& Qwen3-VL-4B   & \underline{0.70}$^\dagger$ & 0.61/0.54 & 0.65/\underline{0.74}$^\ddagger$ & \underline{0.69/0.60} & \underline{0.88/0.82} & 0.73/0.69 \\
& \textbf{TunnelMIND} & \textbf{0.71} & \textbf{0.68/0.67} & \textbf{0.76/0.81}$^\ddagger$ & \textbf{0.76/0.69} & \textbf{0.88/0.90} & \textbf{0.74/0.71} \\
\bottomrule
\end{tabular}
\label{tab:4}
\end{table*}

At the same time, the results also clarify the scope of this transferability. TunnelMIND performs slightly worse than Rex-Omni on worker pose recognition, suggesting that the proposed framework is better suited to defect-, damage-, and region-oriented tasks than to pose-like problems that rely more heavily on dedicated structural modeling. This is consistent with the design focus of TunnelMIND: its main advantage lies in converting open-vocabulary proposals into reliable spatial entities that can support segmentation, measurement, and engineering interpretation. Therefore, the cross-task results should be understood primarily as evidence of interface-level reusability rather than as a claim of universal superiority across all tunnel vision tasks.

\subsection{Entity-level engineering evaluation}\label{sec4:entity}
While the previous sections evaluate whether TunnelMIND improves training-free inspection reliability at the perception level, the central claim of this work is more specific: the framework should convert coarse training-free outputs into measurable and traceable engineering entities. Therefore, the entity-level evaluation in this section serves as the primary validation of engineering usability, rather than as an auxiliary extension to conventional detection metrics.

Because the central claim of TunnelMIND is not merely improved localization but more measurable and traceable engineering interpretation, entity-level evaluation serves as the primary validation of engineering usability. Instead of assessing only whether a target can be roughly localized, this part evaluates whether the predicted results can support downstream engineering quantification, including structural location assignment, geometric measurement, and severity grading under rule-based criteria.

\begin{table}
\centering
\setlength{\heavyrulewidth}{1.2pt}
\setlength{\lightrulewidth}{0.6pt}
\caption{Entity-level engineering evaluation. Lower values indicate better performance for LMAE (length mean absolute error), WMAE (width mean absolute error), and ARE (area relative error), while higher values indicate better performance for LocAcc (location accuracy), SM-F1 (severity macro-averaged F1 score), and EC (percentage of complete entities). $\dagger$ denotes SAM combined with box prompt, and Supervised denotes the task-specific trained reference model.}
\begin{tabular}{lcccccc}
\toprule
\textbf{Method} & \textbf{LMAE (cm)} & \textbf{WMAE (mm)} & \textbf{ARE (\%)} & \textbf{LocAcc (\%)} & \textbf{SM-F1} & \textbf{EC (\%)} \\
\midrule
Qwen3-VL-4B$^\dagger$ & 19.6 & 3.42 & 18.8 & 72.1 & 0.61 & 38.7 \\
Rex-Omni$^\dagger$    & 16.7 & 3.28 & 15.2 & 74.9 & 0.66 & 45.1 \\
\textbf{TunnelMIND}   & \textbf{11.4} & \textbf{1.91} & \textbf{10.3} & \textbf{80.6} & \textbf{0.74} & \textbf{61.8} \\
Supervised            & 6.7 & 0.98 & 7.6 & 93.8 & 0.81 & 75.4 \\
\bottomrule
\end{tabular}
\label{tab:5}
\end{table}

The results in Table~\ref{tab:5} show that TunnelMIND achieves the best performance among all training-free methods. Compared with direct Qwen3-VL-4B$^\dagger$, TunnelMIND reduces the length mean absolute error from 19.6 cm to 11.4 cm, the width mean absolute error from 3.42 mm to 1.91 mm, and the area relative error from 18.8\% to 10.3\%. These improvements indicate that the recalibrated prompts and refined masks provide more accurate geometric support for downstream measurement.

TunnelMIND also improves structural localization and severity assessment, reaching 80.6\% in location accuracy and 0.74 in severity Macro-F1. More importantly, entity completeness increases from 38.7\% for Qwen3-VL-4B$^\dagger$ to 61.8\%, which suggests that TunnelMIND is more effective at producing structured entities with category, location, geometry, and severity information, rather than only coarse proposals or partial segmentation results.

These results are important because the value of TunnelMIND is not exhausted by better visual overlap or higher F1. What matters more in engineering deployment is whether a predicted defect can be turned into a localized, measurable, and gradeable inspection record. From this perspective, the reduction in geometric error and the increase in entity completeness suggest that the proposed intermediate layer between semantic proposals and engineering entities is not only visually helpful, but also practically meaningful for downstream tunnel inspection workflows.

\subsection{Retrieval-grounded report quality}
While Section~\ref{sec4:entity} evaluates whether the predicted results can support measurable and gradeable engineering entities, this section examines whether such structured entities also provide a better interface for retrieval-grounded explanation and report generation. This evaluation is intended to provide a comparative signal of reporting utility rather than to replace formal field-level engineering assessment. Specifically, we investigate whether the proposed entity-centric representation improves retrieval relevance and report quality in a constrained knowledge-grounded setting, where the generated outputs are expected to remain useful, safe, and clear for engineering reading. \textit{The detailed retrieval settings, relevant-fragment definition, and report-scoring rubric are described in Supplementary C.}

\begin{table}
\centering
\setlength{\heavyrulewidth}{1.2pt}
\setlength{\lightrulewidth}{0.6pt}
\caption{Retrieval-grounded report-quality evaluation. VisQuan (visual quantification) uses quantified visual perception results as input, while StrEntity (structured entity) uses reconstructed engineering entities. Hit rate@3 measures whether at least one pre-specified relevant knowledge fragment is retrieved within the top three results. Usefulness, Safety, and Clarity are comparative report-quality scores assigned under a unified rubric, with higher values indicating better retrieval-grounded reporting utility.}
\begin{tabular}{lcccc}
\toprule
\textbf{Method} & \textbf{Hit rate@3} & \textbf{Usefulness} & \textbf{Safety} & \textbf{Clarity} \\
\midrule
VisQuan + LLM       & N/A  & 3.1 & 2.7 & 3.4 \\
VisQuan + RAG       & 71.3 & 3.6 & 3.8 & 3.5 \\
StrEntity + LLM     & N/A  & 3.4 & 3.2 & 3.6 \\
StrEntity + RAG     & 79.6 & 4.0 & 4.2 & 3.9 \\
\textbf{TunnelMIND} & \textbf{84.7} & \textbf{4.3} & \textbf{4.7} & \textbf{4.1} \\
\bottomrule
\end{tabular}
\label{tab:6}
\end{table}

Table~\ref{tab:6} reports the results. Compared with directly using raw visual outputs or less structured descriptions, TunnelMIND achieves the best overall performance in both retrieval and reporting quality. In particular, it obtains the highest Hit@3, indicating that the structured entity formulation improves alignment between detected cases and relevant knowledge fragments in the database. It also achieves the best average scores in usefulness, safety, and clarity, suggesting that reports grounded in category, location, geometry, and severity attributes are more suitable for engineering-oriented explanation than outputs based only on coarse localization or free-form visual interpretation.

\begin{figure}
    \centering
    \includegraphics[width=0.8\linewidth]{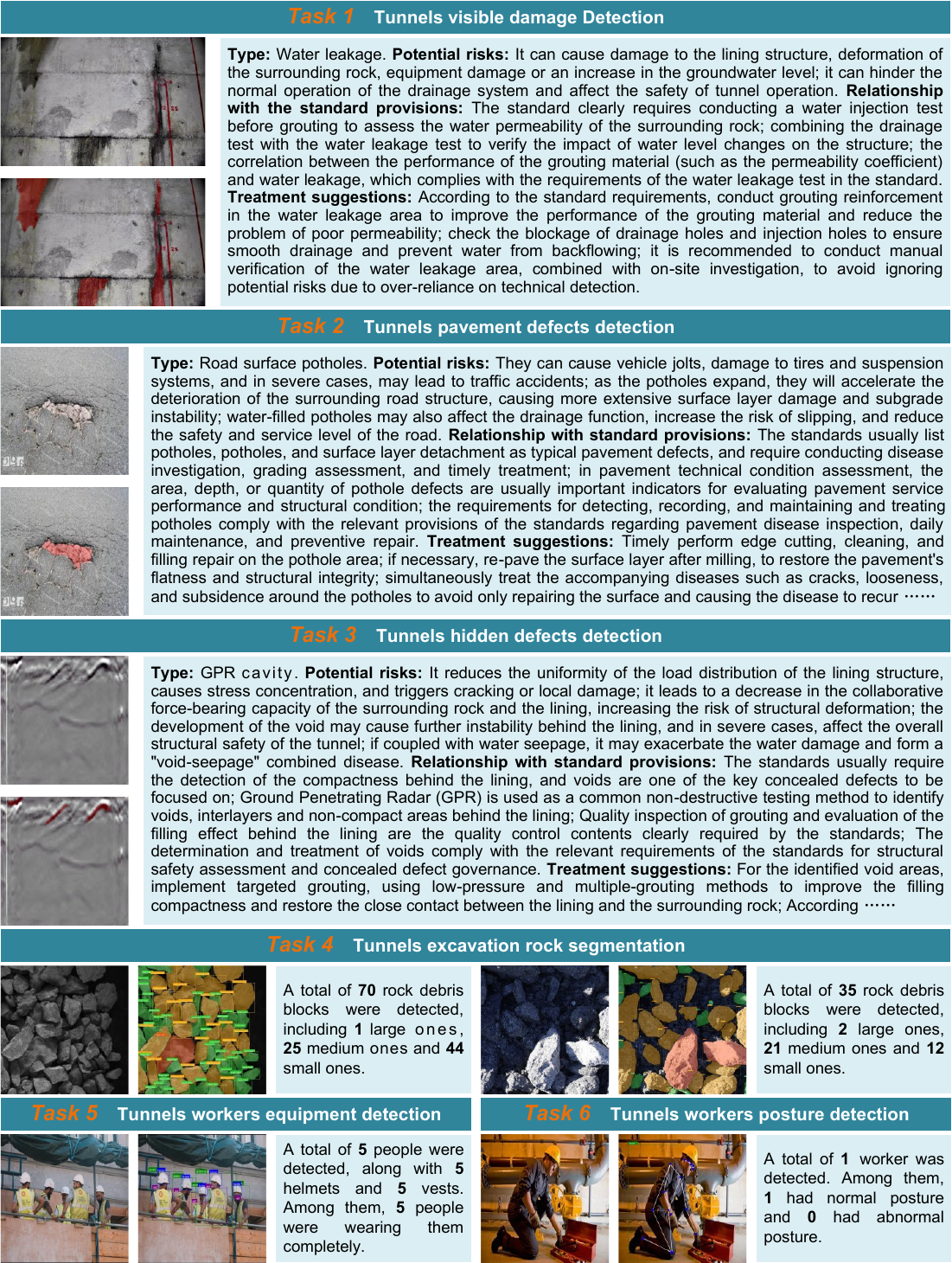}
    \caption{\textbf{Representative end-to-end outputs of TunnelMIND on six tunnel-related tasks.} For visible lining defects, pavement defects, and GPR hidden defects, TunnelMIND provides defect visualization together with engineering-oriented explanations and treatment suggestions. For excavation rock, worker equipment, and worker posture tasks, the framework produces task-specific visual results and concise structured summaries, including rock-block statistics, PPE completeness, and posture status. The figure highlights the deployment-facing capability of TunnelMIND to connect training-free tunnel inspection, entity-centric interpretation, and engineering-readable reporting within a unified framework.}
    \label{fig:9}
\end{figure}

These results should be interpreted with appropriate caution. First, the current evaluation is comparative rather than field-certified. The report scores are based on a model-assisted judging protocol with manual verification, and are intended to reflect relative differences in retrieval-grounded reporting quality across methods. Second, the role of the downstream module is not to automate engineering decision-making, but to provide traceable explanation and advisory report assistance under explicit knowledge constraints. Under this formulation, the main value of TunnelMIND lies in showing that structured entities serve as a more effective interface between training-free perception and engineering-readable reporting.

Beyond the quantitative evaluation, Figure~\ref{fig:9} provides representative end-to-end examples of TunnelMIND on six tunnel-related tasks. The figure shows that the proposed framework can produce task-adaptive visual outputs under a unified inference pipeline, while further converting defect-oriented results into engineering-readable reports when structured entities and retrieval evidence are available. For non-defect tasks such as excavation rock, PPE inspection, and worker posture recognition, the framework returns concise summary-style outputs instead of detailed defect-handling suggestions.

Overall, the results support the claim that entity-centric reconstruction contributes not only to perception-side measurability, but also to downstream interpretability and report quality. This further justifies the design of TunnelMIND as an inspection-oriented pipeline that connects semantic proposal generation, spatial recalibration, engineering entity construction, and retrieval-grounded explanation within a unified training-free framework.

\textbf{Qualitative comparison of engineering effort.} To complement the quantitative evaluation above, we further compare the engineering effort required to onboard a newly introduced inspection task under different deployment paradigms. This comparison focuses on practical task-onboarding cost rather than only inference-time efficiency. Specifically, we consider the amount of new data preparation, annotation workload, setup/debugging time, personnel involvement, the overall time required before a method becomes usable, and its potential for cross-task reuse. From this perspective, the practical value of TunnelMIND lies not only in training-free perception itself, but also in substantially reducing the engineering effort needed to deploy a new tunnel inspection task.

\begin{table}
    \centering
    \caption{\textbf{The comparison reflects task-onboarding effort rather than inference-time efficiency.} Compared with a supervised task-specific pipeline, TunnelMIND requires far fewer new images, no task-specific annotation, substantially less setup/debugging time, fewer personnel, and a much shorter time to usable deployment, while offering higher cross-task reuse. These values are intended as engineering effort estimates under a standardized task-onboarding protocol, rather than audited project logs.}
    \begin{tabular}{lcccccc}
    \toprule[1.2pt]
    Method & New images & Annotation & Training/debugging & Personnel & Time to usable & Cross-task \\
    & to prepare & (person-hours) & (hours) & (persons) & deployment (hours) & reuse \\
    \midrule
    Supervised & 1500 & $\sim$75 & $\sim$36 & 3 & $\sim$120 & Low \\
    TunnelMIND & 20 & 0 & $\sim$2 & 1 & $\sim$4 & High \\
    \bottomrule[1.2pt]
    \label{tab:deployment_effort}
    \end{tabular}
\end{table}

As shown in Table \ref{tab:deployment_effort}, the main engineering advantage of TunnelMIND lies in its much lower task-onboarding burden relative to a supervised task-specific pipeline. Under the estimated setup considered here, the supervised pipeline requires about 1,500 new images, around 75 person-hours of annotation, about 36 hours of training/debugging, and 3 personnel, resulting in an overall onboarding time of about 120 hours. In contrast, TunnelMIND requires only about 20 new images, no task-specific annotation, about 2 hours of lightweight setup/debugging, and 1 person, reducing the time to usable deployment to about 4 hours. This comparison highlights that the practical value of TunnelMIND is not the complete elimination of setup effort, but the substantial reduction of data preparation, annotation, retraining, and coordination cost when a new inspection task must be introduced. Such a difference is particularly relevant to tunnel engineering scenarios in which inspection demands evolve across construction and operation stages and new tasks must be deployed under limited data and manpower.

\subsection{Ablation studies}
To examine the contribution of each component in TunnelMIND, we conduct an ablation study centered on the visual recalibration stage, as reported in Table~\ref{tab:7}. Starting from the raw Qwen3-VL-4B outputs, the F1 scores on Visible, GPR, and Road are 0.57, 0.69, and 0.64, respectively. Adding SAM with point prompts already improves all three tasks, indicating that mapping large-model outputs to pixel-level masks is necessary, although the gain remains limited because the spatial support still mainly depends on the original semantic cues.

\begin{table}
\centering
\setlength{\heavyrulewidth}{1.2pt}
\setlength{\lightrulewidth}{0.6pt}
\caption{Ablation of the recalibration stage. `Q. retrieval' refers to the retrieval of query anchor points in the DINOv3 feature map. `Prototype' denotes the feature prototype obtained through neighborhood clustering after querying. `Neg. points' represent exclusion points incorporated during SAM segmentation. `NMS' stands for post-processing after non-maximum suppression.}
\begin{tabular}{lccc}
\toprule
\textbf{Variant} & \textbf{Visible F1} & \textbf{GPR F1} & \textbf{Road F1} \\
\midrule
Qwen3-VL-4B   & 0.57 & 0.69 & 0.64 \\
+ SAM (point) & 0.62 & 0.70 & 0.67 \\
+ Q. retrieval& 0.65 & 0.73 & 0.67 \\
+ Prototype   & 0.65 & 0.76 & 0.69 \\
+ Neg. points & 0.67 & 0.76 & 0.71 \\
+ NMS         & 0.68 & 0.78 & 0.72 \\
\midrule
\textbf{TunnelMIND} & \textbf{0.68} & \textbf{0.78} & \textbf{0.72} \\
\bottomrule
\end{tabular}
\label{tab:7}
\end{table}

When query-anchor retrieval in the DINOv3 feature space is introduced, the performance on Visible and GPR further improves, showing that dense visual consistency helps correct coarse semantic proposals. Adding the local prototype aggregation strategy brings additional gains, especially on GPR and Road, which suggests that neighborhood-based feature averaging can reduce anchor noise and retrieval drift. These results confirm that the improvement does not come from segmentation alone, but from transferring semantic anchors into a more reliable dense visual support space.

\begin{figure}
    \centering
    \includegraphics[width=0.8\linewidth]{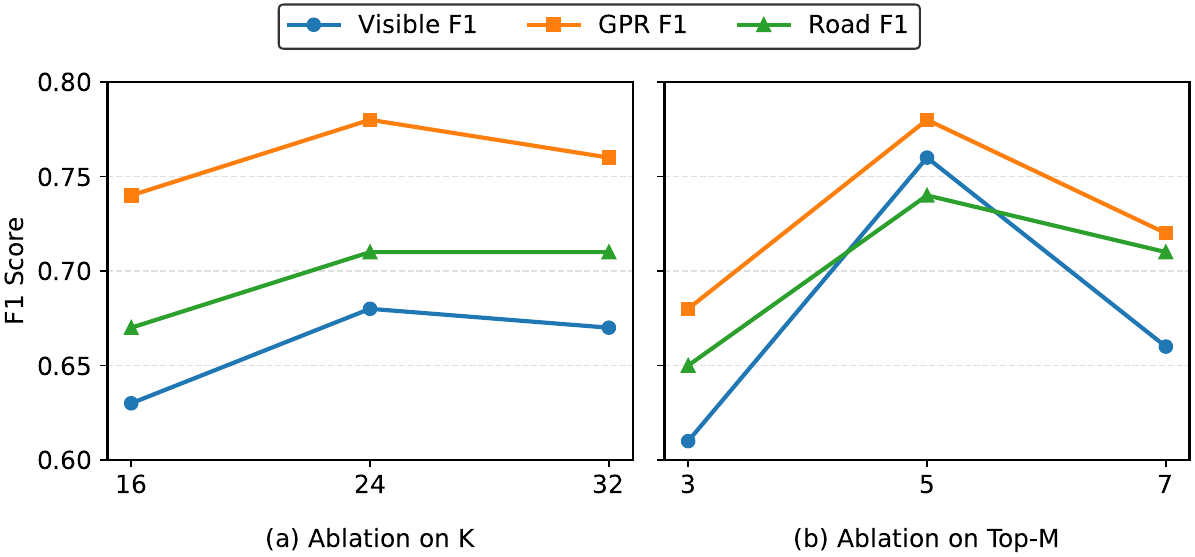}
    \caption{\textbf{Hyperparameter sensitivity of the recalibration stage.} (a) shows the effect of the grid number $K$, and (b) shows the effect of the maximum retained prompt number $\mathrm{Top}\text{-}M$. Results on the Visible, GPR, and Road tasks show that TunnelMIND performs stably within a moderate hyperparameter range and achieves its best overall trade-off around $K=24$ and $\mathrm{Top}\text{-}M=5$.}
    \label{fig:10}
\end{figure}

After negative points are incorporated, the Visible and Road tasks improve again, indicating that explicit exclusion cues can effectively reduce background leakage and over-segmentation in cluttered scenes. Finally, after applying NMS, the three tasks reach their best results, namely 0.68, 0.78, and 0.72. This shows that duplicate suppression remains necessary in multi-candidate cases. Overall, Table~\ref{tab:7} demonstrates that the performance gain of TunnelMIND is not produced by any single component, but by the progressive interaction of point-based segmentation, dense retrieval, prototype aggregation, negative prompting, and post-processing.

Figure~\ref{fig:10} further presents the sensitivity analysis of the key hyperparameters in the recalibration stage. As the grid number $K$ increases from 16 to 24, the F1 scores on all three tasks improve, whereas a further increase to 32 leads to a slight decline. This suggests that overly coarse partitioning limits the precision of local visual retrieval, while overly fine partitioning may introduce additional noise and unstable responses. A similar trend is observed for $\mathrm{Top}\text{-}M$, where $M=5$ yields the best overall performance. When $M$ is too small, the number of positive prompts is insufficient to form stable spatial support; when $M$ is too large, more background interference is introduced, weakening prompt discriminability. Therefore, we adopt $K=24$ and $\mathrm{Top}\text{-}M=5$ as the default configuration. These results indicate that TunnelMIND is reasonably stable within a moderate hyperparameter range, although a balance still needs to be maintained between prompt sparsity and background robustness. \textit{Additional ablation interpretation, runtime breakdown, and deployment analysis are provided in Supplementary E.}

\section{Discussion}\label{sec5}
The results indicate that TunnelMIND provides a practical path from training-free visual grounding to engineering-oriented tunnel inspection. The main contribution of the framework is not simply that it combines a vision-language model, DINOv3, SAM, and retrieval-grounded generation within one pipeline, but that it introduces an explicit intermediate mechanism between semantic discovery and engineering interpretation. In tunnel scenes, language-guided proposals are often semantically plausible but spatially unstable, especially under hard negatives such as shadows, joints, stains, reflections, and repetitive textures. The cross-model visual recalibration module addresses this problem by transferring coarse semantic anchors into a dense visual support space and constraining them with local visual consistency before segmentation and measurement. The representative-task and hard-negative results jointly show that this intermediate step is necessary if training-free perception is expected to move beyond coarse open-vocabulary grounding toward more usable inspection outputs. 

A second implication of this work is that engineering usability depends not only on whether a target can be roughly localized, but also on whether the result can be converted into measurable, traceable, and engineering-compatible structured entities. In TunnelMIND, this is achieved by combining recalibrated segmentation with entity-centric engineering reconstruction, so that category, structural location, geometric attributes, and severity can be organized into a unified representation. This design is particularly valuable for defect- and region-oriented tasks, where downstream interpretation depends directly on length, area, position, and grading information. It also explains why the framework shows greater value on visible defects, GPR hidden defects, and road defects than on pose-like tasks that rely more heavily on dedicated structural modeling. From this perspective, the contribution of TunnelMIND is not merely improved perception under a training-free setting, but the establishment of a usable bridge between semantic proposals and engineering records. 

The retrieval-grounded reporting component should be understood within this same boundary. Its role is not to automate engineering decision-making, but to provide a deployment-facing interface that improves the readability, traceability, and knowledge alignment of inspection outputs. By using structured entities rather than raw boxes, masks, or free-form visual descriptions as retrieval inputs, the reporting module is better able to align defect observations with standards, manuals, and prior cases. The experimental results on report quality support this interpretation, but they should still be viewed as comparative evidence of reporting utility rather than as a substitute for formal engineering review. In other words, the reporting stage extends the practical value of the perception results, but it does not redefine the framework as an autonomous diagnosis system. 

At the same time, the current framework still has several limitations. First, TunnelMIND is not well suited to heavily overlapping, intersecting, or visually entangled defects. In such cases, anchor-centered retrieval and prompt-based segmentation may merge adjacent instances or fail to preserve fine-grained separation, which in turn affects subsequent geometric quantification and severity assessment. Second, the current GPR evaluation follows a fixed visual prompting protocol with one reference exemplar per category. Although this design improves comparability across baselines in weak-semantic radar imagery, it does not fully characterize sensitivity to exemplar choice. Third, practical deployment still requires nontrivial computational resources because the system depends on multiple frozen foundation models executed in sequence. Under the current hardware setting, the perception stages can be completed within a few seconds per image, whereas downstream report generation contributes a large fraction of the total latency. Therefore, the present framework is better suited to offline inspection assistance, report drafting, and assisted review than to strict real-time control. These limitations do not invalidate the proposed design, but they clarify the current deployment boundary of training-free tunnel inspection. 

Overall, TunnelMIND should be viewed as a flexible and deployment-oriented alternative for data-scarce, defect-oriented tunnel inspection scenarios, rather than as a universal replacement for task-specific supervised models. Its main value lies in showing that training-free perception becomes substantially more useful when semantic proposals are complemented by dense visual recalibration and structured engineering reconstruction. Future work will focus on more robust instance disentanglement under overlapping-defect conditions, more systematic analysis of exemplar sensitivity in weak-semantic modalities such as GPR, and lightweight acceleration strategies for improving deployment efficiency.

\section{Conclusion}\label{sec6}
This paper presented TunnelMIND, a training-free tunnel inspection framework that integrates language-guided semantic anchoring, cross-model visual recalibration, prompt-based geometric realization, and entity-centric engineering reconstruction within a unified inspection pipeline. Instead of treating a vision-language model as a final detector, TunnelMIND uses semantic proposals as intermediate anchors and refines their spatial support in the dense DINOv3 feature space, so that visually consistent evidence can be converted into more reliable prompts for segmentation and downstream quantification. The resulting masks are then organized into structured engineering entities and further linked to retrieval-grounded explanation and report assistance. 

Experiments on representative tunnel inspection tasks show that the proposed framework improves the usability of training-free perception, remains more robust under hard-negative conditions, and produces higher-quality engineering entities for downstream interpretation. These findings suggest that the main obstacle in training-free tunnel inspection is not only whether a model can respond to open-vocabulary task descriptions, but whether its outputs can be transformed into stable and measurable engineering results. TunnelMIND addresses this obstacle by introducing an explicit intermediate layer between semantic candidate generation and engineering interpretation. 

Although challenges remain in overlapping-defect separation, exemplar sensitivity in weak-semantic modalities, and deployment efficiency, the present study demonstrates that training-free perception can move beyond coarse localization toward more traceable and engineering-compatible tunnel inspection outputs. In this sense, TunnelMIND provides a practical basis for data-scarce, defect-oriented tunnel inspection and deployment-facing retrieval-grounded inspection assistance. 

\section{Acknowledgment}

\section{Declaration of generative AI and AI-assisted technologies in the manuscript preparation process.}
During the preparation of this work the authors used GPT-5.4 for translation and polishing. After using this tool/service, the authors reviewed and edited the content as needed and takes full responsibility for the content of the published article.

\clearpage 





\bibliographystyle{cas-model2-names}

\bibliography{cas-refs}



\end{document}